\documentclass{bmvc2k}
\usepackage{graphicx}
\usepackage{wrapfig}
\usepackage{amsmath,amssymb,amsfonts}
\usepackage{tabularx}
\usepackage{booktabs}
\usepackage{multirow}
\usepackage{xspace}
\usepackage{microtype}
\usepackage{algorithm}
\usepackage{algorithmic}
\usepackage{float}
\usepackage{appendix}
\usepackage{multirow}
\usepackage{pifont}
\usepackage{xcolor}

\newcommand{\cmark}{\ding{51}}
\newcommand{\xmark}{\ding{55}}

\newcommand{\method}{GaussVLA\xspace}

\title{GaussVLA: Geometry-Aware Spatial Reasoning for Vision-Language-Action Model}

\addauthor{Md Selim Sarowar}{selim.sarowar12@gmail.com}{1}
\addauthor{Md Tanvir Islam}{tanvirnwu@knu.ac.kr}{2}
\addauthor{Sungho Kim}{sunghokim@yu.ac.kr}{$1^*$}
\addauthor{Sangtae Ahn}{stahn@knu.ac.kr}{2}

\addinstitution{
 Yeungnam University,\\
 Republic of Korea
}
\addinstitution{
 Kyungpook National University,\\
 Republic of Korea
}

\runninghead{Md Selim Sarowar et al.}{GaussVLA}

\begin{document}

\maketitle
\let\thefootnote\relax\footnote{$^*$Corresponding author. \href{https://gaussvla.github.io/GaussVLA/}{Project page: GaussVLA}}
\vspace{-0.5cm}
\begin{abstract}
Vision-Language-Action (VLA) models encode visual observations as flat 2D patch tokens that carry no intrinsic geometric structure, and augmenting them with dense monocular depth injects per-pixel scalar values that encode neither surface orientation nor geometric confidence. This leaves the policy with limited structured spatial reasoning for action prediction. We propose \emph{GaussVLA}, a mamba-based VLA that incorporates two custom modules: \emph{Gaussian Spatial Tokenizer (GST)} to lift frozen semantic and depth features into compact 3D Gaussian tokens, pools geometrically salient regions with learned queries, and \emph{Depth-Aware Chain-of-Thought (DA-CoT)} that performs structured, non-autoregressive geometric reasoning under language and flow-time conditioning. Across both simulation and real-world evaluations, \emph{GaussVLA} demonstrates strong spatial-manipulation performance while remaining parameter efficient. On LIBERO, it achieves 93.5\% average success and 100.0\% success on the Spatial suite with only 200M parameters, improving over SpatialVLA by 19.7\% relative average success while remaining significantly more parameter-efficient. 
\end{abstract}


\section{Introduction}
\label{sec:intro}

\begin{wrapfigure}{r}{0.4\textwidth}
  \centering 
  \vspace{-12mm}
  \includegraphics[width=1.0\linewidth]{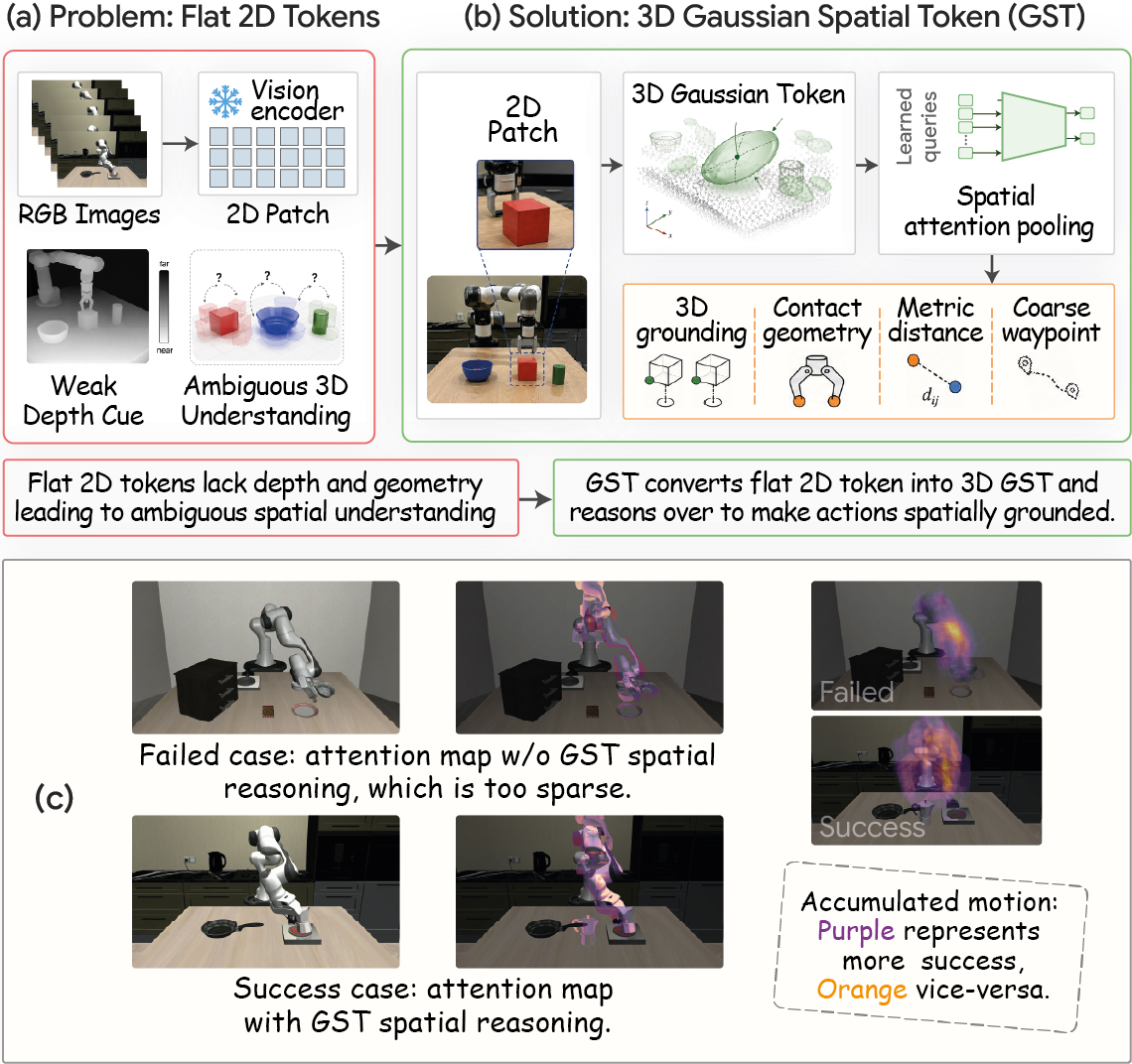}
  \vspace{-7mm}
  \caption{\small Overview of \emph{GaussVLA}.}
  \label{fig:firingRate}
  \vspace{-5mm}
\end{wrapfigure}
Learning robot manipulation policies from demonstrations has become a dominant paradigm, where vision-language-action (VLA) models~\cite{pmlr-v270-kim25c,pmlr-v229-zitkovich23a,black2025tpo} leverage visual observations and language conditioning to enable generalizable robotic skills across diverse tasks. These models inherit strong language grounding from pretrained vision-language backbones~\cite{marafioti2025smolvlm,zhai2023sigmoid} and translate this grounding into executable, language-conditioned manipulation policies. However, their visual front-end projects each RGB observation into a flat sequence of 2D patch tokens. While these tokens capture appearance and semantic information, they lack explicit metric 3D geometry and spatial information, treating all patches equally regardless of depth, surface orientation, or geometric reliability. Although such representations can achieve good performance on standard benchmarks, they become less reliable when tasks require precise geometric reasoning and robustness to variations.

Recent work attempts to address this limitation by injecting geometric information, either by augmenting visual tokens with monocular depth estimates~\cite{qu2025spatialvla} or by incorporating additional reasoning mechanisms into the policy pipeline~\cite{tur2026recurrentdepth}. While effective to some extent, these approaches typically model depth as a per-pixel scalar signal, which primarily encodes distance but does not explicitly capture surface orientation or geometric reliability. As a result, important spatial cues, such as local surface structure, must be inferred implicitly from depth patterns, which can be ambiguous. Moreover, treating all depth predictions as equally reliable ignores the estimator's confidence that varies across regions~\cite{ 10.5555/3295222.3295309}.

Beyond representation, current VLA pipelines remain limited in their ability to reason about spatial relationships prior to action decoding. While modern flow-matching and diffusion-based action heads~\cite{black2025tpo,li2024cogact} generate continuous action sequences conditioned on backbone features, spatial relationships are encoded only implicitly, without modeling of geometric structure. Autoregressive Chain-of-Thought (CoT) reasoning~\cite{zawalski2024robotic,zhao2025cotvla} introduces intermediate reasoning steps, but operates primarily in $2D$ image space, incurs latency from sequential token generation, and remains decoupled from action decoding without direct spatial supervision.

To address these limitations, we propose a model \emph{GaussVLA} that integrates structured \emph{GST} with reasoning-aware action decoding. Instead of representing visual observations as flat $2D$ tokens or scalar depth maps, \emph{GST} encodes the observations as a set of $3D$ Gaussian primitives that capture spatial position, local geometry, and depth-aware features in a continuous representation, thereby enabling the modeling of geometric relationships. Building on this, we introduce \emph{DA-CoT}, which injects structured spatial reasoning directly into the action decoding process through supervised multimodal tokens, aligning reasoning with action generation. To efficiently model the resulting structured, long-range dependencies in our proposed \emph{GaussVLA}, we adopt a mamba-based backbone~\cite{gu2024mamba}, which enables linear-time sequence modeling and improved computational efficiency over attention-based architectures. Together, these components enable improved spatial reasoning and more reliable performance in robotic manipulation tasks, marked by the following key contributions:
\begin{itemize}
    \item \textbf{Gaussian Spatial Tokenizer (GST).} We introduce a structured 3D spatial representation that converts depth and visual features into a compact set of Gaussian primitives. Each primitive is parameterized by a 3D mean that captures spatial position, a covariance that encodes local geometric structure and spatial extent, and an opacity term that represents depth-aware confidence, enabling structured modeling of scene geometry. The resulting representation is refined through spatial attention pooling.

    \item \textbf{Depth-Aware Chain-of-Thought (DA-CoT).} We introduce a structured supervision mechanism for multimodal token prediction that incorporates spatial reasoning signals into the action generation process. \emph{DA-CoT} introduces a structured, non-autoregressive reasoning-conditioning module in which a small set of learnable queries cross-attends to \emph{GST} tokens under joint language and flow-time conditioning, thereby anchoring the reasoning representation to the action-prediction objective.
    
    \item \textbf{Efficient unified framework.} We present a unified VLA framework \emph{GaussVLA} that integrates \emph{GST} and \emph{DA-CoT} within an efficient sequence modeling backbone, enabling real-time inference with a relatively low computational footprint. We benchmark our approach on four widely used manipulation datasets, achieving performance comparable to state-of-the-art baselines while being more computationally efficient.
\end{itemize}

\section{Related Works}
\label{sec:related}

Previous VLAs~\cite{pmlr-v270-kim25c,rt12022, ZitkovichYXXXXW23,zheng2025tracevla,pmlr-v229-zitkovich23a} discretize action dimensions with BPE-style binning and fine-tune pretrained VLMs to autoregressively emit action tokens. While this route preserves language grounding, it ties the plan to long sequences that accumulate prediction errors over time, leading to a drop in performance and a high computation cost as chunk length increases.

On the other hand, a prominent line of research reformulates action generation as a continuous denoising process, drawing directly on Diffusion Policy~\cite{chi2025diffusion}. By modeling the visuomotor policy as a conditional Denoising Diffusion Probabilistic Model (DDPM), Diffusion Policy excels at capturing multimodal action distributions and producing temporally smooth, high-dimensional trajectories. However, Diffusion Policy~\cite{chi2025diffusion} requires multiple denoising steps per action, which slows execution, and has no language grounding.

Later, this paradigm has been widely integrated into VLA architectures by appending a dedicated diffusion action head to the vision-language backbone. For instance, models such as Octo~\cite{octo_2023} and TinyVLA~\cite{wen2025tinyvla} leverage this decoupled architecture, utilizing a transformer or a lightweight VLM followed by a diffusion head to generate continuous action chunks. This approach successfully bypasses the precision limits of discrete binning while ensuring stable, multimodal control. But Octo~\cite{octo_2023} requires large pre-training data and struggles to process egocentric views, which makes it underperform. Furthermore, TinyVLA requires multiple sampling steps, which sacrifices performance and semantic understanding.

To overcome this trade-off between these paradigms, more recent systems further iterate on decoupling action decoding from the autoregressive backbone. $\pi_0$~\cite{black2025tpo} attaches a flow-matching action expert to a VLM via a mixture-of-transformers with shared attention, recovering continuous chunked prediction at the cost of an additional head. CogACT~\cite{li2024cogact} uses a diffusion head with classifier-free guidance, which smooths trajectories but adds tens of denoising steps per chunk. To mitigate the inference latency of these generative heads, $\pi_0$-FAST~\cite{pertsch2025fast} replaces per-step tokenization with a frequency domain code (Discrete Cosine Transform) that compresses action chunks before autoregressive decoding, trading expressivity for length. PD-VLA~\cite{11247519} reformulates autoregressive decoding as a nonlinear system solved by parallel fixed-point iterations, recovering wall-clock speed at the price of a bounded number of refinement steps. 

\vspace{0.25cm}
\noindent
\textbf{Spatial Awareness and Visual Reasoning.}
Meanwhile, recent works such as 3D-VLA~\cite{ZhenQCY0DHG24} incorporate RGB-D observations and point-cloud representations to make VLA policies more spatially aware.
Similarly, SpatialVLA~\cite{qu2025spatialvla} injects pseudo point-cloud tokens into the policy. However, such representations can lose fine-grained object morphology and remain sensitive to depth-estimation errors and sensor noise. Therefore, effectively bridging these geometric priors to low-level control demands structured reasoning mechanisms, which introduce a fundamental trade-off.

In parallel, CoT-VLA~\cite{zhao2025cotvla} and ECoT~\cite{zawalski2024robotic} perform textual reasoning in 2D image coordinates before action prediction; this is interpretable but incurs autoregressive latency on the order of seconds per step. RD-VLA~\cite{tur2026recurrentdepth} moves reasoning into a latent recurrent space to recover speed, but sacrifices the auditability of chain-of-thought and remains constrained by bounded depth generalization, as excessive recurrent unrolling leads to state saturation and degraded performance.

Beyond efficiency, action representation strongly affects generalization. QueST~\cite{mete2024quest} uses finite scalar quantization to learn a task-agnostic vocabulary of action primitives for few-shot transfer, while Mask2Act~\cite{Zhang_2025_BMVC} learns latent plans from action-free video by predicting future object masks instead of full RGB frames. However, its reliance on 2D masks limits access to the 3D geometry needed for precise, contact-rich manipulation.

\vspace{0.25cm}
\noindent
\textbf{SSM in Robot Manipulation.} Consequently, to process rich 3D sequences and structured reasoning traces without prohibitive latency, the underlying model architecture must be highly efficient. Standard transformer VLA backbones incur quadratic attention cost over the joint multimodal sequence and maintain a KV cache that grows with the chunk horizon. Selective state-space models (SSMs) such as Mamba~\cite{gu2024mamba} address this by replacing attention with a linear-time selective scan and a constant-size hidden state. This is highly attractive for sequences that mix language, visual patches, 3D tokens, and streaming proprioception. Recent work has demonstrated the viability of replacing transformer stacks with SSM cores to achieve favorable throughput at matched representational capacity. For instance, MaIL~\cite{jia2024mail} applies this successfully to imitation learning, while RoboMamba~\cite{liurobomamba} leverages the Mamba architecture to efficiently co-train vision-language reasoning and continuous motor control. However, the open question for an SSM-based VLA is how to retain random access to spatial geometry, since the linear scan summarizes context into a bounded state and cannot natively re-read individual primitives.

In this study, we introduce \emph{GaussVLA} to geometrically address the issues by pushing residual correction, confidence estimation, and depth-aware CoT-derived gradients directly into the tokenizer. It resolves the latency-interpretability trade-off by supervising a short, structured CoT in metric 3D coordinates rather than relying on an open-ended textual trace. \emph{GaussVLA} resolves this by coupling the Mamba core to the full $P{=}256$ Gaussian field through a global cross-attention mechanism. By mapping learned reasoning queries to pooled \emph{GST} tokens. It preserves the linear-time property of the scan while enabling the reasoning path to query fine-grained 3D primitives on demand.

\section{Proposed Method: GaussVLA}
\label{sec:method}

We propose \textbf{GaussVLA}, an end-to-end VLA framework that equips robotic policy learning with structured 3D spatial representation and reasoning-aware action generation. Given a window of RGB observations and a language instruction, \emph{GaussVLA} first performs dual-stream visual encoding to extract complementary semantic and geometric cues. These two streams are then fused by our \emph{Gaussian Spatial Tokenization (GST)} module, which transforms dense 2D patch tokens into structured 3D Gaussian spatial tokens that encode semantic content, spatial location, and confidence-aware geometric structure. Building on these tokens, we introduce \emph{Depth-Aware Chain-of-Thought} (DA-CoT), which aggregates task-relevant spatial relations into a small set of reasoning tokens and injects them into a Mamba-based multimodal backbone. Finally, a flow-matching policy head predicts a conditional velocity field over action trajectories, thereby generating the target action sequence. Thus, \emph{GaussVLA} combines three complementary ingredients within a unified architecture: geometry-aware representation learning, reasoning-aware action conditioning, and efficient linear-time sequence modeling.

\begin{figure}[t]
    \centering
    \includegraphics[width=\textwidth]{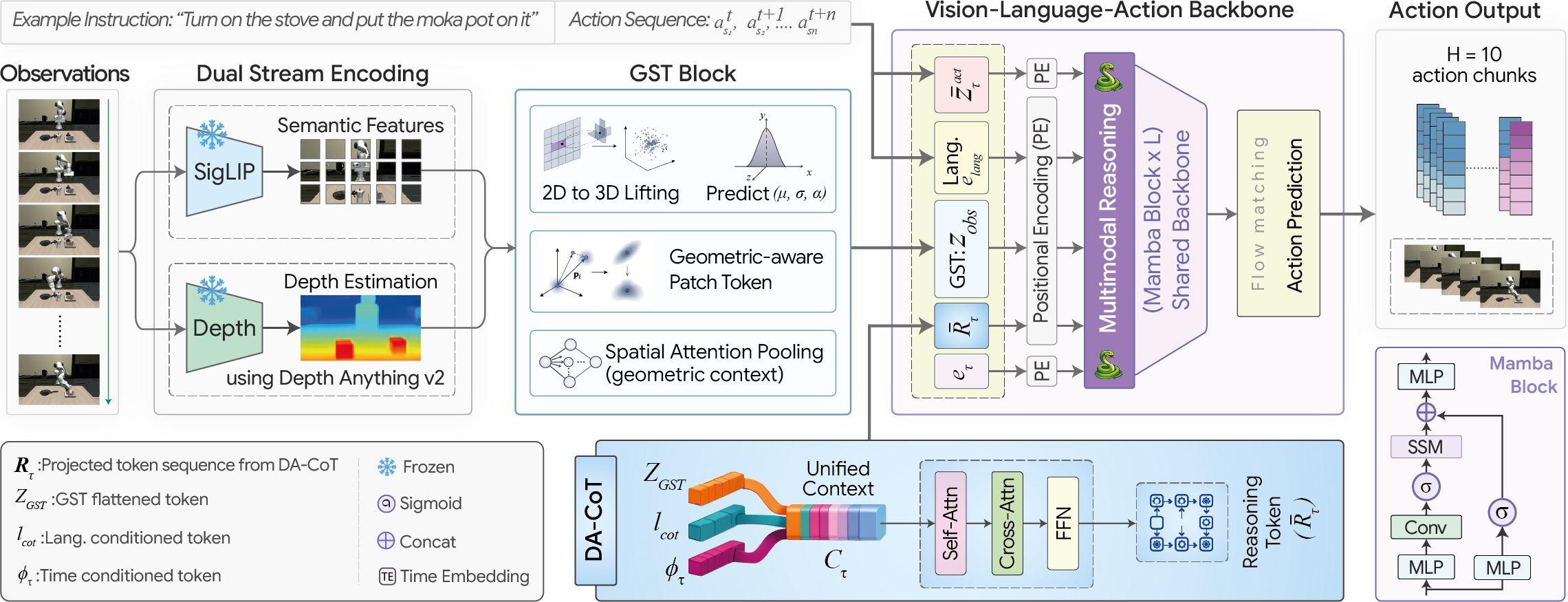}
    \vspace{-0.5cm}
    \caption{\small \textbf{Framework of our proposed \emph{GaussVLA} for VLA-based Robot Manipulation.} A frozen dual-stream encoder (SigLIP and Depth Anything V2) extracts complementary semantic and dense depth features for geometric context. These streams are fused by the Gaussian Spatial Tokenizer (GST), which lifts 2D image evidence into structured 3D Gaussian tokens via Gaussian parameters $(\mu, \sigma, \alpha)$, forming geometry-aware patch tokens, and then aggregates them through spatial attention pooling. The resulting \emph{GST} tokens are combined with language, CoT, action, and temporal embeddings, and processed by a shared Mamba-based multimodal backbone with positional encoding. Finally, a \emph{DA-CoT} layer refines spatial reasoning before the flow-matching action prediction head.}
\label{fig:GaussVLA_overview}
\vspace{-0.25cm}
\end{figure}

\subsection{Problem Formulation}
\label{sec:problem_formulation}
We study language-conditioned robot manipulation from demonstrations. The training set is
\(\mathcal{D}=\{(\mathcal{O}^{(n)}, s^{(n)}, \mathbf{A}^{(n)})\}_{n=1}^{N}\), where \(N\) is the number of training samples. For each sample, \(\mathcal{O}=\{\mathbf{x}_{t}^{(m)}\}_{t=1,m=1}^{T_o, M}\) denotes a window of \(T_o\) RGB observations from \(M\) camera views, where \(\mathbf{x}_{t}^{(m)} \in \mathbb{R}^{3 \times H_x \times W_x}\) is the image captured at observation step \(t\) from camera \(m\), and \(H_x\) and \(W_x\) are the image height and width. The language instruction is denoted by \(s\), and its encoded representation is given by $\mathbf{l}=E_{\mathrm{lan}}(s)\in\mathbb{R}^{d_l}$.
where $E_{\mathrm{lan}}$ is the language encoder and $d_l$ is the language embedding dimension. The expert action trajectory is $\mathbf{A}=[\mathbf{a}_1,\ldots,\mathbf{a}_H]\in\mathbb{R}^{H \times d_a}$, where $H$ is the action horizon, $d_a$ is the action dimension, and $\mathbf{a}_t \in \mathbb{R}^{d_a}$ is the action at step $t$.

Our goal is to learn a policy \(\pi_{\theta}\), parameterized by \(\theta\), that predicts a future action sequence \(\hat{\mathbf{A}}=\pi_{\theta}(\mathcal{O},\mathbf{l})\). In \emph{GaussVLA}, this prediction is obtained through a structured pipeline: RGB observations are first converted into semantic and geometric features, then compressed into 3D Gaussian spatial tokens, refined with depth-aware reasoning tokens, and finally decoded into actions by a Mamba-based \cite{gu2024mamba} backbone with a flow-matching head.

In \emph{GaussVLA}, we learn a conditional velocity field for action generation in order to sample Gaussian noise \(\boldsymbol{\epsilon}\sim\mathcal{N}(\mathbf{0},\mathbf{I})\), 
and a flow time \(\tau \in [0,1]\) from a distribution \(p(\tau)\). We then form the interpolated action state \(\mathbf{Z}_{\tau}=(1-\tau)\mathbf{A}+\tau\boldsymbol{\epsilon}\), whose analytic target velocity is \(\mathbf{U}=\boldsymbol{\epsilon}-\mathbf{A}\). Conditioned on \(\mathbf{Z}_{\tau}\), the observation context \(\mathcal{O}\), the language embedding \(\mathbf{l}\), and the flow time \(\tau\), the model predicts a velocity field \(v_{\theta}(\mathbf{Z}_{\tau},\mathcal{O},\mathbf{l},\tau)\). 

%
The following subsections describe how \emph{GaussVLA} constructs the structured representation through dual-stream visual encoding, \emph{GST}, and \emph{DA-CoT} reasoning.

\subsection{Dual-Stream Visual Encoding}
Conventional VLA policies typically encode each RGB observation into 2D visual tokens, leaving the geometric structure to be implicitly inferred by the downstream policy. In contrast, \emph{GaussVLA} decomposes visual perception into two complementary streams before spatial tokenization: a \emph{semantic stream} for appearance-aware patch features and a \emph{geometric stream} for depth-aware spatial cues.

For each image $\mathbf{x}_{t}^{(m)} \in \mathcal{O}$, we apply two frozen encoders: a semantic encoder \cite{zhai2023sigmoid} and a depth encoder \cite{depth_anything_v2}.
The semantic stream uses a pretrained patch encoder $E_{\mathrm{sem}}$ to extract a set of patch-level visual features as follows:
\begin{equation}
\mathbf{F}_{t}^{(m)} = E_{\mathrm{sem}}(\mathbf{x}_{t}^{(m)}) \in \mathbb{R}^{P \times d_s},
\label{eq:semantic_stream}
\end{equation}
where $P$ is the number of image patches and $d_s$ is the semantic feature dimension. We denote the feature of patch $r \in \{1,\dots,P\}$ by $\mathbf{f}_{t,r}^{(m)} \in \mathbb{R}^{d_s}$. These patch features retain local appearance information and high-level semantic cues useful for distinguishing objects, parts, and task-relevant visual patterns.

In parallel, the depth stream uses a pretrained monocular depth encoder $E_{\mathrm{dep}}$ to estimate depth from the same RGB input. To align this stream with the patch representation in Eq.~\eqref{eq:semantic_stream}, we sample the predicted depth map at the centers of the $P$ image patches as follows:
\begin{equation}
\mathbf{d}_{t}^{(m)} = E_{\mathrm{dep}}(\mathbf{x}_{t}^{(m)})\downarrow_{\mathrm{patch}} \in \mathbb{R}^{P},
\label{eq:geometric_stream}
\end{equation}
where $\downarrow_{\mathrm{patch}}$ denotes patch-center sampling, and $d_{t,r}^{(m)} \in \mathbb{R}$ is the depth value associated with patch $r$. This produces a patch-aligned semantic-depth representation in which each patch is described by the pair $(\mathbf{f}_{t,r}^{(m)}, d_{t,r}^{(m)})$.

\subsection{Gaussian Spatial Tokenization (GST)}
Given the patch-aligned semantic $(\mathbf{f}_{t,r}^{(m)})$ and depth features $(d_{t,r}^{(m)})$ from the dual-stream encoder, the goal of \emph{GST} is to convert dense 2D patch tokens into structured 3D Gaussian tokens using its patch-center image coordinate $\mathbf{u}_r = [u_r, v_r, 1]^\top$ and predicted depth. To compensate for monocular depth scale bias, we apply a learnable affine correction and back-project it as follows:
\begin{equation}
\tilde d_{t,r}^{(m)} = \lambda_d\, d_{t,r}^{(m)} + \beta_d,
\label{eq:learable_affine}
\end{equation}
\begin{equation}
\mathbf{c}_{t,r}^{(m)} = \tilde d_{t,r}^{(m)} \bigl(\mathbf{K}^{(m)}\bigr)^{-1} \mathbf{u}_r \in \mathbb{R}^3,
\label{eq:gst_backprojection}
\end{equation}
where $\mathbf{K}^{(m)} \in \mathbb{R}^{3\times 3}$ is the camera intrinsic matrix; $\lambda_d, \beta_d \in \mathbb{R}$ are learnable scalars.

Starting from the semantic feature \(\mathbf{f}_{t,r}^{(m)}\), a learnable Gaussian parameter head \(f_{\mathrm{par}}\) predicts a residual mean offset \(\Delta \boldsymbol{\mu}_{t,r}^{(m)} \in \mathbb{R}^{3}\), a log-scale vector \(\boldsymbol{\eta}_{t,r}^{(m)} \in \mathbb{R}^{3}\), and a confidence score \(\alpha_{t,r}^{(m)} \in (0,1)\) as follows:

\begin{equation}
\left(
\Delta \boldsymbol{\mu}_{t,r}^{(m)},
\boldsymbol{\eta}_{t,r}^{(m)},
\alpha_{t,r}^{(m)}
\right)
=
f_{\mathrm{par}}\!\left(\mathbf{f}_{t,r}^{(m)}\right).
\label{eq:gst_param_head}
\end{equation}
\noindent\textit{Scope of $\alpha$.}
The opacity $\alpha_{t,r}^{(m)}$ is predicted from the semantic feature $\mathbf{f}_{t,r}^{(m)}$ alone and does not directly consume a depth-uncertainty signal. Depth-awareness arises \emph{implicitly} through the opacity-weighted GST loss (Eq.~\eqref{eq:gst_loss}): high-$\alpha$ patches whose back-projection disagrees with the corrected depth incur a larger penalty, inducing low $\alpha$ in geometrically ambiguous regions. We verify this induced correlation in the Supplementary Material \textcolor{red}{A.09}.

We then define the Gaussian mean as follows:
\begin{equation}
\boldsymbol{\mu}_{t,r}^{(m)}
=
\mathbf{c}_{t,r}^{(m)}
+
\rho \tanh\!\left(\Delta \boldsymbol{\mu}_{t,r}^{(m)}\right),
\label{eq:gst_mean}
\end{equation}
where \(\rho > 0\) is a fixed workspace radius that bounds the residual offset. The corresponding Gaussian scale (log-variance) is as follows:
\begin{equation}
\boldsymbol{\sigma}_{t,r}^{(m)} = \exp\!\left(\boldsymbol{\eta}_{t,r}^{(m)}\right).
\label{eq:gst_scale}
\end{equation}
Thus, each patch gives rise to a Gaussian primitive
mean \(\big(\boldsymbol{\mu}_{t,r}^{(m)})\), log-variance \((\boldsymbol{\sigma}_{t,r}^{(m)})\), where \(\alpha_{t,r}^{(m)}\) reflects the reliability of the local geometry.

To give it 3D geometry awareness, we apply a 3D Fourier positional encoding to each Gaussian mean as follows:
\begin{equation}
\gamma\!\left(\boldsymbol{\mu}_{t,r}^{(m)}\right)
=
\Big[
\sin\!\left(2^0 \pi \boldsymbol{\mu}_{t,r}^{(m)}\right),
\cos\!\left(2^0 \pi \boldsymbol{\mu}_{t,r}^{(m)}\right),
\dots,
\sin\!\left(2^{L-1} \pi \boldsymbol{\mu}_{t,r}^{(m)}\right),
\cos\!\left(2^{L-1} \pi \boldsymbol{\mu}_{t,r}^{(m)}\right)
\Big],
\label{eq:gst_fourier}
\end{equation}

where the sine and cosine functions are applied element-wise to the three spatial coordinates, yielding
$\gamma(\boldsymbol{\mu}_{t,r}^{(m)}) \in \mathbb{R}^{6L}$.

We then form the geometry-aware patch token as follows:
\begin{equation}
\label{eq:geo}
\mathbf{g}_{t,r}^{(m)}
=
\left[
\mathbf{f}_{t,r}^{(m)}
\oplus
\gamma\!\left(\boldsymbol{\mu}_{t,r}^{(m)}\right)
\oplus
\boldsymbol{\eta}_{t,r}^{(m)}
\right]
\in \mathbb{R}^{d_g},
\quad
d_g = d_s + 6L + 3,
\end{equation}
where the $\oplus$ denotes concatenation along the token dimension.

The token $\mathbf{g}_{t,r}^{(m)}$ from Eq.~\eqref{eq:geo} jointly encodes semantic appearance, 3D position, and local Gaussian geometry through the log-scale vector
$\boldsymbol{\eta}_{t,r}^{(m)}$. The confidence term $\alpha_{t,r}^{(m)}$ is used separately as a reliability bias in the subsequent spatial attention pooling step.

We first project the patch tokens into a pooling space using 
\(\tilde G_{t}^{(m)} = G_{t}^{(m)} \mathbf{W}_{\mathrm{in}}\), 
where \(\mathbf{W}_{\mathrm{in}} \in \mathbb{R}^{d_g \times d_p}\) is a learnable projection matrix and \(d_p\) is the pooling dimension. The pooling module maintains \(N_g\) learnable spatial queries, collected in a matrix \(Q \in \mathbb{R}^{N_g \times d_p}\). For clarity, we write the pooling step in a single-head form, while the actual implementation uses multi-head cross-attention. The pooled attention output is as follows:
\begin{equation}
\hat Z_t^{(m)} = \mathrm{softmax}\!\left(\tfrac{(\mathrm{LN}_1(Q)\mathbf{W}_Q)(\tilde G_t^{(m)}\mathbf{W}_K)^\top}{\sqrt{d_p}} + \mathbf{1}_{N_g}\log(\boldsymbol{\alpha}_t^{(m)})^\top\right)(\tilde G_t^{(m)}\mathbf{W}_V),
\label{eq:gst_pooling}
\end{equation}
where $\mathbf{1}_{N_g}\in\mathbb{R}^{N_g}$ broadcasts the confidence bias across queries. A standard residual FFN block yields the final \emph{GST} features $Z_t^{(m)} = \mathrm{LN}_3(\mathrm{FFN}(\mathrm{LN}_2(Q + \hat Z_t^{(m)})))\in\mathbb{R}^{N_g\times d_z}$. We denote the \(j\)-th row of \(Z_{t}^{(m)}\) by \(\mathbf{z}_{t,j}^{(m)} \in \mathbb{R}^{d_z}\), for \(j \in \{1,\dots,N_g\}\).

Collecting the tokens over all observation steps and camera views yields the structured \emph{GST} representation as follows:
\begin{equation}
\mathcal{Z}_{\mathrm{GST}}
=
\left\{
\mathbf{z}_{t,j}^{(m)}
\right\}_{t=1, m=1, j=1}^{T_o, M, N_g},
\label{eq:gst_representation}
\end{equation}
which serves as the compact geometry-aware visual representation passed to our subsequent \emph{DA-CoT} reasoning module and the action decoder.

\subsection{Depth-Aware Chain-of-Thought (DA-CoT)}
While \emph{GST} provides a compact set of geometry-aware visual tokens, these tokens are still purely perceptual. To extract task-relevant spatial relations before action decoding, we introduce a \emph{DA-CoT} conditioning module. The key idea is to use a small set of learnable reasoning queries to attend over the \emph{GST} representation under joint language and flow-time conditioning, thereby producing structured reasoning tokens.

We begin by flattening the \emph{GST} representation in Eq.~\eqref{eq:gst_representation} across observation steps, camera views, and Gaussian tokens as follows:
\begin{equation}
\bar Z_{\mathrm{GST}}
=
\mathrm{Flatten}\!\left(\mathcal{Z}_{\mathrm{GST}}\right)
\in
\mathbb{R}^{(T_o M N_g)\times d_z},
\label{eq:dacot_flatten}
\end{equation}
where \(d_z\) is the \emph{GST} token dimension. We then project these tokens into a reasoning space of dimension \(d_c\) using a learnable mapping \(f_{\mathrm{gst}}^{\mathrm{cot}}\). In parallel, the language embedding \(\mathbf{l}\) from the section~\ref{sec:problem_formulation} is projected to a language-conditioning token
\(\mathbf{l}_{\mathrm{cot}} = f_{\mathrm{lang}}^{\mathrm{cot}}(\mathbf{l}) \in \mathbb{R}^{d_c}\),
and the flow time \(\tau\) is encoded into a time-conditioning token
\(\boldsymbol{\phi}_{\tau} = f_{\mathrm{time}}(\tau) \in \mathbb{R}^{d_c}\).
These components are concatenated to form the \emph{DA-CoT} context as follows:
\begin{equation}
\mathcal{C}_{\tau}
=
\left[
f_{\mathrm{gst}}^{\mathrm{cot}}(\bar Z_{\mathrm{GST}})\oplus
\mathbf{l}_{\mathrm{cot}}^{\top}\oplus
\boldsymbol{\phi}_{\tau}^{\top}
\right]
\in
\mathbb{R}^{(T_o M N_g + 2)\times d_c}.
\label{eq:dacot_context}
\end{equation}

Next, \emph{DA-CoT} maintains \(N_r\) learnable reasoning queries, collected in a matrix
\(Y^{(0)} \in \mathbb{R}^{N_r \times d_c}\), where \(N_r \ll T_o M N_g\). These queries are refined through self-attention, cross-attention to the context \(\mathcal{C}_{\tau}\), and a feed-forward update as follows:
\begin{align}
Y^{(1)}
&=
\mathrm{LN}_1
\!\left(
Y^{(0)} + \mathrm{MHSA}(Y^{(0)})
\right),
\label{eq:dacot_selfattn}
\\
Y^{(2)}
&=
\mathrm{LN}_2
\!\left(
Y^{(1)} + \mathrm{MHCA}(Y^{(1)}, \mathcal{C}_{\tau}, \mathcal{C}_{\tau})
\right),
\label{eq:dacot_crossattn}
\\
R_{\tau}
&=
\mathrm{LN}_3
\!\left(
Y^{(2)} + \mathrm{FFN}(Y^{(2)})
\right)
\in
\mathbb{R}^{N_r \times d_c},
\label{eq:dacot_tokens}
\end{align}
where \(\mathrm{MHSA}(\cdot)\) denotes multi-head self-attention, \(\mathrm{MHCA}(\cdot,\cdot,\cdot)\) denotes multi-head cross-attention, and \(\mathrm{LN}_1\), \(\mathrm{LN}_2\), and \(\mathrm{LN}_3\) are layer-normalization operators. The rows of \(R_{\tau}\) are the final \emph{DA-CoT} tokens.

Each \emph{DA-CoT} token summarizes a distinct subset of the \emph{GST} representation, conditioned on task and flow time. Intuitively, these tokens provide a structured, non-autoregressive summary of the 3D scene's reasoning, allowing the policy to focus on the spatial relations most relevant to the current manipulation objective, as further presented in the following section.

\subsection{GaussVLA}
This stage integrates the \emph{GST} observation tokens from Eq.~\eqref{eq:gst_pooling}, the \emph{DA-CoT} reasoning tokens from Eq.~\eqref{eq:dacot_tokens}, language conditioning, and flow-state action tokens into a unified policy backbone. Its goal is to parameterize the conditional velocity field for flow-matching action generation. For clarity, we omit the batch dimension.

We first map all inputs to the common backbone dimension $d_b$. Specifically, the \emph{GST} representation $\mathbf{Z}_{\mathrm{GST}}$ from Eq.~\eqref{eq:gst_representation} is flattened over observation steps, camera views, and Gaussian queries, then projected to
$\bar{\mathbf{Z}}_{\mathrm{obs}}\in\mathbb{R}^{T_{\mathrm{GST}}\times d_b}$, where
$T_{\mathrm{GST}}=T_oMN_g$. The \emph{DA-CoT} tokens from Eq.~(16) are projected to
$\bar{\mathbf{R}}_\tau\in\mathbb{R}^{N_r\times d_b}$. The language embedding, flow time, and interpolated action state from Sec .~\ref {sec:problem_formulation} are embedded as
$\mathbf{e}_{\mathrm{lang}}=f_{\mathrm{lang}}(\mathbf{l})$,
$\mathbf{e}_{\tau}=f_{\tau}(\tau)$, and
$\bar{\mathbf{Z}}_{\tau}^{\mathrm{act}}=f_{\mathrm{act}}(\mathbf{Z}_{\tau})$.
The resulting backbone tokens are concatenated along the token dimension to form the multimodal policy input as follows:
\begin{equation}
\mathbf{Y}_\tau
= f_{\mathrm{mamba}}
\left(
\mathbf{e}_\tau
\oplus
\mathbf{e}_{\mathrm{lang}}
\oplus
\bar{\mathbf{Z}}_{\mathrm{obs}}
\oplus
\bar{\mathbf{R}}_\tau
\oplus
\bar{\mathbf{Z}}_{\tau}^{\mathrm{act}}
\right)
\in
\mathbb{R}^{(2+T_{\mathrm{GST}}+N_r+H)\times d_b}.
\end{equation}


Let $\mathbf{Y}^{\mathrm{act}}_\tau\in\mathbb{R}^{H\times d_b}$ denote the final $H$ backbone states corresponding to the action tokens, and let $\mathbf{Y}^{\mathrm{ctx}}_\tau$ denote the remaining non-action states. The action decoder refines the action states using the contextual states, \emph{DA-CoT} tokens, and time embedding:
\begin{equation}
\tilde{\mathbf{Y}}^{\mathrm{act}}_\tau
=
f_{\mathrm{dec}}
\left(
\mathbf{Y}^{\mathrm{act}}_\tau,
\mathbf{Y}^{\mathrm{ctx}}_\tau,
\bar{\mathbf{R}}_\tau,
\mathbf{e}_\tau
\right)
\in
\mathbb{R}^{H\times d_b}.
\label{eq:action_decoder_hidden}
\end{equation}
Finally, the velocity head predicts the conditional flow velocity as follows:
\begin{equation}
\mathbf{v}_\theta
=
f_{\mathrm{vel}}
\left(
\tilde{\mathbf{Y}}^{\mathrm{act}}_\tau
\right)
\in
\mathbb{R}^{H\times d_a}.
\label{eq:velocity_head}
\end{equation}

Thus, \emph{GaussVLA} aligns geometry-aware observation tokens, \emph{DA-CoT} reasoning, and flow-state action tokens within a single conditional policy backbone.

\subsection{Training Objectives}
\emph{GaussVLA} is trained with three complementary objectives: the primary
flow-matching loss, a \emph{GST} depth-consistency loss, and a \emph{DA-CoT} supervision loss. Together, these terms encourage accurate action generation, geometrically consistent spatial tokens, and reasoning features that remain aligned with the target velocity field. For our proposed \emph{GaussVLA}, the flow matching loss is defined as follows:
\begin{equation}
\mathcal{L}_{\mathrm{FM}} = \mathbb{E}_{(\mathcal{O}, s, \mathbf{A}) \sim \mathcal{D}, \, \tau \sim p(\tau), \, \boldsymbol{\epsilon} \sim \mathcal{N}(\mathbf{0}, \mathbf{I})}
\left[
\left\|
v_{\theta}(\mathbf{Z}_{\tau}, \mathcal{O}, \mathbf{l}, \tau) - \mathbf{U}
\right\|_2^2
\right].
\label{eq:flow_matching_loss}
\end{equation}
where $v_\theta$ is predicted velocity and \(\mathbf{U}\) is target velocity as defined in section~\ref{sec:problem_formulation}.



To preserve geometric consistency within \emph{GST}, we use an opacity-weighted depth reconstruction loss on the $z$-coordinate $\mu_{t,r,z}^{(m)}$ of each Gaussian mean (Eq.~\eqref{eq:gst_param_head}) against the affine-corrected depth $\tilde d_{t,r}^{(m)}$ (Eq.~\eqref{eq:learable_affine}), with $\alpha_{t,r}^{(m)}$ the patch confidence as follows:
\begin{equation}
\mathcal{L}_{\mathrm{GST}} = \frac{\sum_{t,m,r}\alpha_{t,r}^{(m)}\,\bigl|\mu_{t,r,z}^{(m)} - \tilde d_{t,r}^{(m)}\bigr|}{\sum_{t,m,r}\alpha_{t,r}^{(m)}}.
\label{eq:gst_loss}
\end{equation}


We attach an auxiliary supervision term so that the \emph{DA-CoT} tokens remain predictive of
the target flow velocity. Let
$\bar{R}_\tau = [\mathbf{r}_{\tau,1},\ldots,\mathbf{r}_{\tau,N_r}]
\in \mathbb{R}^{N_r \times d_b}$
denote the \emph{DA-CoT} tokens after projection to the backbone dimension, and define their mean summary as follows:
\begin{equation}
\bar{\mathbf{r}}_\tau
=
\frac{1}{N_r}\sum_{j=1}^{N_r}\mathbf{r}_{\tau,j}
\in \mathbb{R}^{d_b}.
\end{equation}
This summary is broadcast across the $H$ action steps and combined with the
action-decoder hidden output $\tilde{Y}^{\mathrm{act}}_\tau \in \mathbb{R}^{H \times d_b}$.
A learnable residual head $f_{\mathrm{CoT}}^{\mathrm{res}}$ then predicts an auxiliary
velocity:
\begin{equation}
\hat{\mathbf{U}}_{\mathrm{CoT}}
=
f_{\mathrm{CoT}}^{\mathrm{res}}
\left(
\tilde{Y}^{\mathrm{act}}_\tau
+
\mathbf{1}_{H}\bar{\mathbf{r}}_\tau^{\top}
\right)
\in \mathbb{R}^{H \times d_a},
\end{equation}
where $\mathbf{1}_{H}\in\mathbb{R}^{H}$ broadcasts the \emph{DA-CoT} summary over the action horizon. Given the flow-matching target velocity
$\mathbf{U}_\tau \in \mathbb{R}^{H \times d_a}$, the \emph{DA-CoT} supervision loss is as follows:
\begin{equation}
\mathcal{L}_{\mathrm{CoT}}
=
\frac{1}{H d_a}
\left\|
\hat{\mathbf{U}}_{\mathrm{CoT}} - \mathbf{U}_\tau
\right\|_F^2 .
\end{equation}

The final training objective is the weighted sum as follows:
\begin{equation}
\mathcal{L} = \mathcal{L}_{\mathrm{FM}} + \mathcal{L}_{\mathrm{GST}} + \mathcal{L}_{\mathrm{CoT}}.
\label{eq:total_loss}
\end{equation}


\section{Experiments}
\label{sec:experiments}

\subsection{Experimental Setup}

\vspace{0.25cm}
\noindent{\textbf{Benchmarks.}}
We evaluate \emph{GaussVLA} across simulation, robustness, long-horizon control, and real-world manipulation. LIBERO~\cite{liu2023libero} serves as the primary simulation benchmark. LIBERO-PRO~\cite{zhou2025libero} tests robustness and Meta-World~\cite{mete2024quest} evaluates difficulty-stratified manipulation across Easy, Mid, Hard, and Very Hard splits, while CALVIN~\cite{9788026} ABCD$\rightarrow$D tests long-horizon sequential generalization on the 10\% subset by measuring 1-5 consecutive task completions. We further evaluate sim-to-real transfer on a physical SO-101 robot using multi-task and Pick-Place in-distribution (ID) or out-of-distribution (OOD) trials.

\vspace{0.25cm}
\noindent{\textbf{Metrics.}}
Report success rate (SR) for LIBERO, Meta-World, and SO-101; normalized success under perturbations for LIBERO-PRO; completed sequence length and 1-5 task completion rates for CALVIN.

\vspace{0.25cm}
\noindent{\textbf{Implementation.}}
We instantiate $E_{\mathrm{sem}}$ as frozen SigLIP-SO400M/14 with $P=256$ patch tokens and feature dimension $d_s=1152$, $E_{\mathrm{dep}}$ as frozen Depth-Anything-V2 (ViT-L), and $E_{\mathrm{lang}}$ as a frozen CLIP text encoder. All encoders remain frozen; only \emph{GST}, \emph{DA-CoT}, the Mamba backbone, and the action decoder are trained. \emph{GST} uses $N_g=128$ Gaussian queries per camera with a workspace radius of $0.8$m, while \emph{DA-CoT} uses $N_r=4$ reasoning queries with 512-dimensional hidden states. \textcolor{black}{\emph{GST} uses $L=10$ Fourier band.} The policy backbone uses a backbone dimension $d_b=512$ with 5 Mamba blocks \textcolor{black}{and intermediate dimension 1024}, and predicts 7-DoF action chunks over a horizon $H=10$ using a flow-matching action decoder with 10 Euler steps. \textcolor{black}{The action decoder uses two cross-attention blocks with 4 heads, and learned positional embeddings are used for the multimodal token sequence.} We train with AdamW, learning rate $2.5\times10^{-5}$, weight decay $10^{-4}$, batch size 256, and BF16 precision. \textcolor{black}{The flow time is sampled from $\mathrm{Beta}(1.5,1.0)$ and the learning rate follows a cosine schedule.} The auxiliary loss weights are set to $\lambda_{\mathrm{GST}}=0.05$ and $\lambda_{\mathrm{CoT}}=0.10$ with a 5-epoch warm-up. \textcolor{black}{Reported parameter counts exclude the frozen external SigLIP, Depth-Anything-V2, and CLIP backbones; Full \emph{GaussVLA} has 200M total model parameters and 179M trainable parameters.} Inference runs at 12.97 ms/step on a single NVIDIA RTX Pro 6000 Blackwell. 
Following the official LIBERO benchmark settings, we performed 40 rollouts for each task.
\textcolor{black}{For real-world SO-101 experiments, we use a 6-DoF arm with a parallel-jaw gripper and a RealSense camera at 640$\times$480 and 60 Hz; control runs at 15 Hz with replanning every $H=10$ steps, using 50 teleoperated demonstrations per task. More details are provided in Supplementary Material.}

\begin{table}[t]
\centering \footnotesize
\renewcommand{\arraystretch}{1.0}
\setlength{\tabcolsep}{8pt}
\begin{tabular}{llcccccc} \hline
Method & Venue & Spatial & Object & Goal & Long & Average & Parameters \\
\hline
DP~\cite{chi2025diffusion} & RSS'23 & 78.3 & 92.5 & 68.3 & 50.5 & 72.4 & 80M \\
MaIL~\cite{jia2024mail} & CoRL'24 & 53.8 & 81.5 & 56.3 & 41.7 & 58.3 & 24M \\
QueST~\cite{mete2024quest} & NeurIPS'24 & 89.0 & 90.0 & 88.4 & 87.0 & 88.6 & 152M \\
OpenVLA~\cite{pmlr-v270-kim25c} & CoRL'24 & 84.7 & 88.4 & 79.2 & 53.7 & 76.5 & 7B \\
SpatialVLA~\cite{qu2025spatialvla} & RSS'25 & 88.2 & 89.9 & 78.6 & 55.5 & 78.1 & 4B \\
ThinkAct~\cite{huang2025thinkact} & NeurIPS'25 & 88.3 & 91.4 & 87.1 & 70.9 & 84.4 & 7B \\
CoT-VLA~\cite{zhao2025cotvla} & CVPR'25 & 81.5 & \underline{91.6} & 87.6 & 69.0 & 82.43 & 7B \\
Mask2Act~\cite{Zhang_2025_BMVC} & BMVC'25 & 76.1 & 68.7 & 75.1 & 30.6 & 62.6 & 7B \\
TraceVLA~\cite{zheng2025tracevla} & ICLR'25 & 84.6 & 85.2 & 75.1 & 54.1 & 74.8 & 4B \\
$\pi_0$~\cite{black2025tpo} & RSS'26 & 90.0 & 86.0 & \underline{95.0} & 73.0 & 86.0 & 3.3B \\
SUREFlow\cite{islam2026sureflow} & IROS'26 & \underline{94.8} & 91.0 & 93.8 & \textbf{90.2} & \underline{92.5} & 179.1M \\
\hline
\textbf{\method} & \textbf{BMVC'26} & \textbf{100} & \textbf{95.8} & \textbf{95.3} & \underline{83.0} & \textbf{93.5} & \textbf{1B} \\
\hline
\end{tabular}
\vspace{0.15cm}
\caption{\small Performance comparison on the LIBERO~\cite{liu2023libero} benchmark. \emph{GaussVLA} achieves the highest overall average success rate while using substantially fewer parameters than large-scale VLA baselines.}
\label{tab:libero_results}
\end{table}
\subsection{Simulation Results}
\noindent\textit{\textbf{LIBERO.}}
Table~\ref{tab:libero_results} summarizes the main simulation results on LIBERO~\cite{liu2023libero}. \emph{GaussVLA} achieves the best overall average success rate of 93.5\%, outperforming the strongest prior average, QueST, by 4.9 points, and substantially exceeding large-scale VLA baselines such as $\pi_0$, OpenVLA, SpatialVLA, ThinkAct, and CoT-VLA. The gains are especially clear on geometry-sensitive suites: \emph{GaussVLA} reaches 100.0\% on Spatial, 95.8\% on Object, and 95.3\% on Goal. 
\begin{wrapfigure}{r}{0.60\textwidth}
  \centering \vspace{-0.25cm}
  \includegraphics[width=1.0\linewidth]{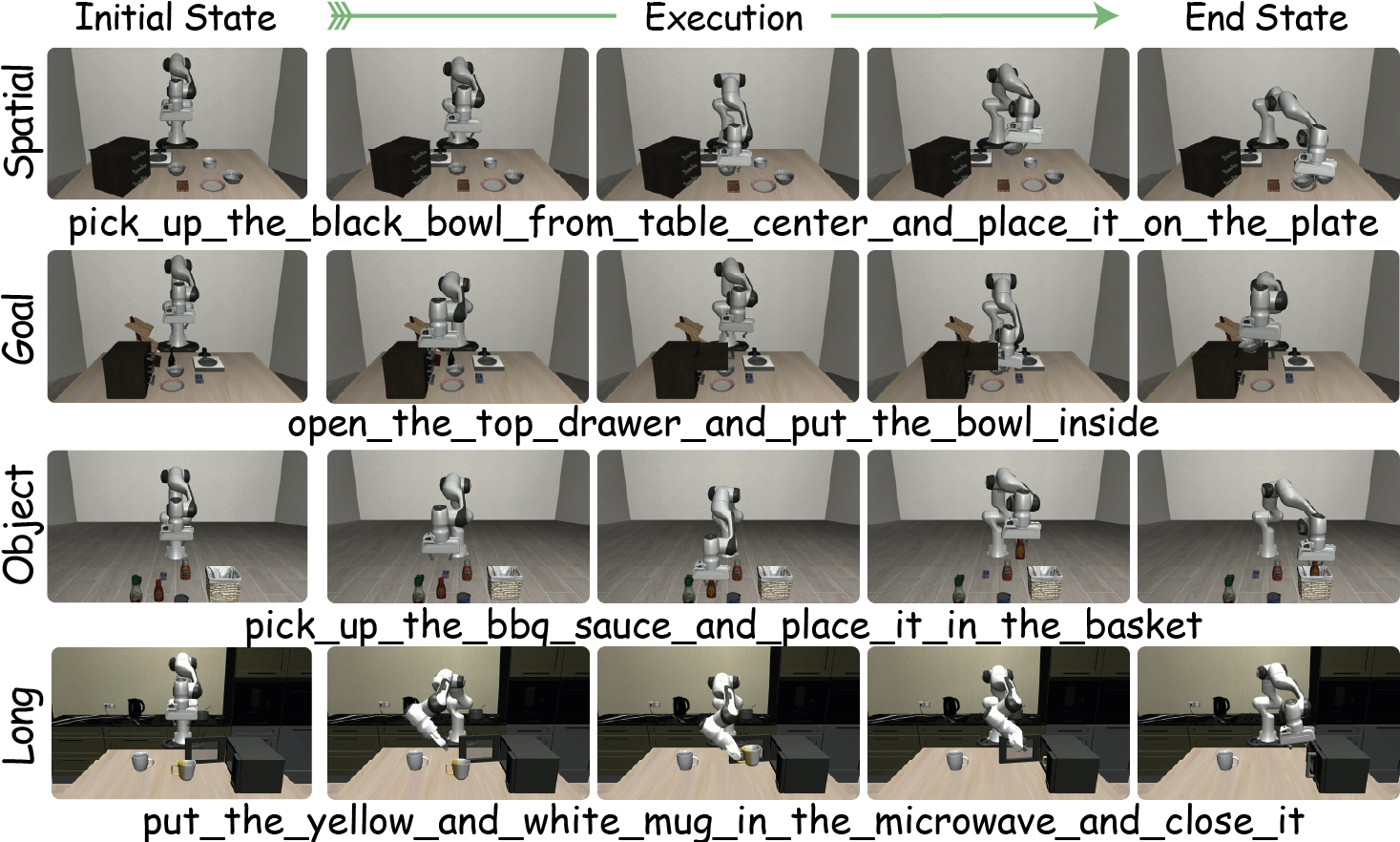}
  \vspace{-0.5cm}
  \caption{\small Representative LIBERO \cite{liu2023libero} rollouts showing \emph{GaussVLA} executing tasks from initial state to completion.} 
  \label{fig:libero_simulation}
  \vspace{-0.25cm}
\end{wrapfigure}
This suggests that the proposed \emph{GST} provides a useful 3D structure rather than merely adding model capacity. On the Long suite, \emph{GaussVLA} obtains 83.0\%, ranking second behind QueST while remaining well above most VLA baselines, indicating that the model preserves temporal consistency across extended manipulation episodes. 
Importantly, these gains are achieved with only 200M trainable parameters, far smaller than several 3.3B-7B VLA systems, highlighting the parameter efficiency of the proposed spatial reasoning design. Representative LIBERO rollouts in Fig.~\ref{fig:libero_simulation} further illustrate successful execution from initial state to goal state across Spatial, Goal, Object, and Long tasks.

\begin{table}[t]
\centering \scriptsize 
\setlength{\tabcolsep}{1.5pt} \renewcommand{\arraystretch}{1.2}
\begin{tabular}{l l c c c c c | llcccccc}
\hline
\multirow{2}{*}{Method} & \multirow{2}{*}{Venue} & \multicolumn{5}{c|}{Meta-World} & \multirow{2}{*}{Method} & \multirow{2}{*}{Venue} & \multicolumn{5}{c}{Tasks completed in a row (CALVIN)} & Avg.\\
\cline{3-7} \cline{10-14}
 &  & Easy & Mid & Hard & V.Hard & Avg. &  &  & 1 & 2 & 3 & 4 & 5 & Length \\
\hline
BC-RNN~\cite{robomimic2021}     & CoRL'21 & 4.5  & 3.8  & 3.2  & 3.0  & 3.6 & MCIL\cite{LynchS21} & RSS'21 & 0.37 & 0.027 & 0.002 & 0.000 & 0.000 & 0.40 \\
DP~\cite{chi2025diffusion}        & IJRR'25 & \underline{83.6} & 31.1 & 9.0  & 26.6 & 37.6 & MT-R3M\cite{nair2022rm} & CoRL'22 & 0.408 & 0.146 & 0.043 & 0.014 & 0.002 & 0.61 \\
TinyVLA~\cite{wen2025tinyvla} & ICRA'25 & 77.6 & 21.5 & 11.4 & 15.8 & 31.6 & RT-1\cite{rt12022} & RSS'23 & 0.249 & 0.069 & 0.015 & 0.006 & 0.000 & 0.34 \\
$\pi_0$~\cite{black2025tpo} & RSS'26 & 71.8 & \textbf{48.2} & \textbf{41.7} & \underline{30.0} & \underline{47.9} & GR-1\cite{bjorck2025gr00t} & ICLR'24 & \textbf{0.778} & \textbf{0.533} & \textbf{0.332} & \textbf{0.218} & \textbf{0.139} & \textbf{2.0} \\

\hline
\textbf{GaussVLA} & \textbf{BMVC'26} & \textbf{92.7} & \underline{47.9} & \underline{37.3} & \textbf{41.6} & \textbf{54.9} &
\textbf{\method} & \textbf{BMVC'26} & \underline{0.637} & \underline{0.367} & \underline{0.291} & \underline{0.179} & 0.000 & \underline{1.474} \\
\hline
\end{tabular}
\vspace{-0.15cm}
\caption{\small Comparison with recent imitation learning and VLA baselines on Meta-World \cite{mclean2025metaworld} and CALVIN \cite{9788026}. The Meta-World panel reports SR across splits by task difficulty. The CALVIN panel reports long-horizon performance as the fraction of rollouts that complete 1-5 consecutive tasks, along with the average length of completed sequences.}
 \label{tab:metaworld_results}
\end{table}

\vspace{1.0cm}
\noindent\textit{\textbf{Meta-World and CALVIN.}}
Table~\ref{tab:metaworld_results} evaluates whether the same design transfers beyond LIBERO~\cite{liu2023libero} to difficulty-stratified manipulation and long-horizon sequential control. On Meta-World, \emph{GaussVLA} achieves the best performance on the Easy split and remains highly competitive as task difficulty increases, obtaining the second-best average performance among the compared methods. A notable trend is that several baselines show strong performance only in limited regimes: DP and TinyVLA perform well on Easy tasks but degrade sharply on Hard and Very Hard settings, whereas \emph{GaussVLA} maintains a more balanced profile across difficulty levels. Although $\pi_0$ leads on the Mid and Hard splits, \emph{GaussVLA} closely matches the best result on Very Hard tasks and achieves a strong overall average, suggesting that its geometric tokenization is particularly useful for manipulation tasks that require robust spatial grounding.

On CALVIN, \emph{GaussVLA} consistently ranks second behind GR-1 for completing one to four consecutive tasks and achieves an average sequence length of 1.474. Compared with MT-R3M, \emph{GaussVLA} improves the 1, 2, 3, and 4-task completion rates by large margins, indicating that the policy not only succeeds at isolated actions but also maintains coherent behavior across multi-step rollouts. The remaining gap to GR-1, especially at five consecutive tasks, indicates that very long-horizon recovery remains challenging; however, \emph{GaussVLA's} strong performance through four-task sequences demonstrates that the proposed depth-aware reasoning improves temporal robustness and reduces early compounding failures. Representative Meta-World simulation frames are provided in the supplementary.

\begin{table}[ht]
\centering \scriptsize
\setlength{\tabcolsep}{1.4pt}
\renewcommand{\arraystretch}{1.0}
\begin{tabular}{lccccc |c ccccc |c ccccc |c ccccc |c}
\hline
\multirow{2}{*}{Model} & \multicolumn{5}{c|}{LIBERO Goal} & &
  \multicolumn{5}{c|}{LIBERO Spatial} & &
  \multicolumn{5}{c|}{LIBERO 10} & &
  \multicolumn{5}{c|}{LIBERO Object} & Avg.\\
\cline{2-6} \cline{8-12} \cline{14-18} \cline{20-24}
& Obj & Pos & Sem & Task & Env & &
  Obj & Pos & Sem & Task & Env & &
  Obj & Pos & Sem & Task & Env & &
  Obj & Pos & Sem & Task & Env & SR $\uparrow$ \\
\hline
OVLA~\cite{pmlr-v270-kim25c}
& 0.96 & 0.00 & 0.98 & 0.00 & 0.98 & &
  0.97 & 0.00 & 0.97 & 0.00 & 0.89 & &
  0.81 & 0.00 & 0.96 & 0.00 & 0.85 & &
  0.98 & 0.00 & 0.98 & 0.00 & 0.00 & \underline{0.52}\\
$\pi_{0.5}$~\cite{black2025pi}
& 0.97 & 0.38 & 0.97 & 0.00 & 0.46 & &
  0.97 & 0.20 & 0.97 & 0.01 & 0.46 & &
  0.92 & 0.08 & 0.93 & 0.01 & 0.46 & &
  0.98 & 0.17 & 0.96 & 0.01 & 0.73 & \textbf{0.53}\\
  
$\pi_{0}$~\cite{black2025tpo}
& 0.94 & 0.00 & 0.93 & 0.00 & 0.39 & &
  0.95 & 0.00 & 0.97 & 0.00 & 0.60 & &
  0.79 & 0.00 & 0.82 & 0.00 & 0.27 & &
  0.94 & 0.00 & 0.90 & 0.00 & 0.29 & 0.44\\
    
\hline
\textbf{\method}
    & 0.81 & 0.00 & 0.70 & 0.00 & 0.00 & &
    0.93 & 0.00 & 0.73 & 0.00 & 0.27 & &
    0.83 & 0.00 & 0.67 & 0.00 & 0.00 & &
    0.92 & 0.00 & 0.63 & 0.00 & 0.17 & 0.33\\ \hline
\end{tabular}
\caption{\small Comparison on LIBERO-PRO \cite{zhou2025libero} benchmark. Normalized success rates (SR) under five perturbation types. Here, \textbf{OVLA}: OpenVLA~\cite{pmlr-v270-kim25c}.}
\label{tab:libero_pro}
\end{table}

\begin{figure}[ht]
    \centering
    \includegraphics[width=\textwidth]{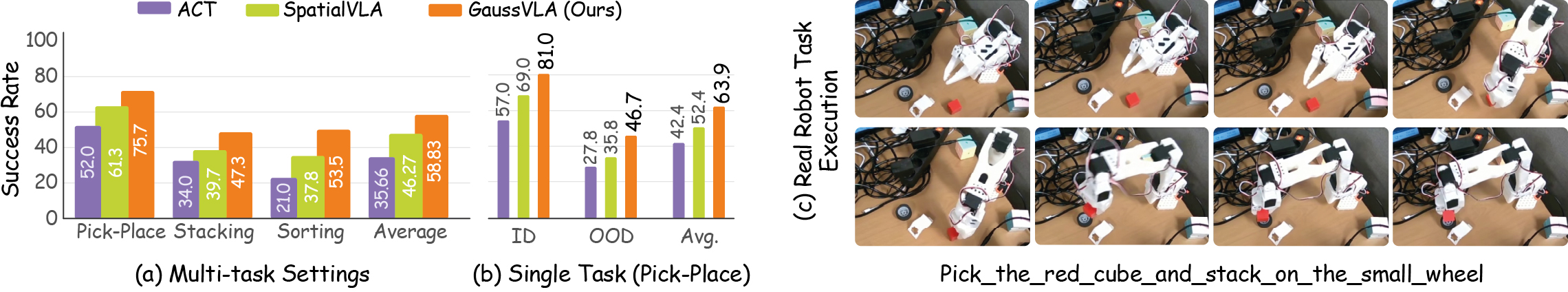}
    \vspace{-0.5cm}
    \caption{\small Real-world SO-101 evaluation comparing \emph{GaussVLA} with ACT and SpatialVLA across multi-task settings and Pick-Place ID and OOD trials, alongside rollout frames showing successful execution of a physical stacking task.}
\label{fig:real_results}
 \vspace{-0.5cm}
\end{figure}

\subsection{Robustness and Real-World Evaluation}

We first evaluate robustness on LIBERO-PRO~\cite{zhou2025libero}, which introduces substantially larger distribution shifts than standard LIBERO by varying object appearance, position, semantics, task instructions, and environments. As shown in Table~\ref{tab:libero_pro}, this benchmark remains highly challenging: even VLA models built on large pretrained VLM backbones suffer severe performance drops, indicating that LIBERO-PRO tests robustness beyond in-distribution imitation. Although \emph{GaussVLA} does not achieve the highest LIBERO-PRO score, it obtains a meaningful average SR of 0.33 with a lightweight 200M-parameter architecture. This suggests that structured Gaussian Spatial Tokenization provides useful transfer under distribution shift without relying on massive model scale. 

Since simulation robustness alone does not fully capture deployment behavior, we further evaluate \emph{GaussVLA} on a real-world SO-101 robot platform. Figure~\ref{fig:real_results} shows that \emph{GaussVLA} achieves the highest success rate across all real-world multi-task categories, improving the average success rate to 58.8\% compared with 46.3\% for SpatialVLA and 35.7\% for ACT. The gain is especially clear on Sorting, where \emph{GaussVLA} outperforms SpatialVLA by 15.7 points, highlighting the benefit of spatial tokenization under real camera noise and \emph{GaussVLA} execution uncertainty. In the Pick-Place setting, GaussVLA reaches 81.0\% ID trials and 46.7\% on OOD trials, outperforming SpatialVLA by 12.0 and 10.9 points, respectively. These results indicate that \emph{GaussVLA} provides a stable geometric basis for manipulation as object placement and scene configuration change.

\begin{table}[ht]
\centering
\scriptsize
\renewcommand{\arraystretch}{1.0}
\setlength{\tabcolsep}{1pt}
\begin{tabular}{lccc|cc|cccc}
\hline
Variant 
& 2D Tokens 
& GST 
& DA-CoT 
& LIBERO $\uparrow$ 
& LIBERO-PRO $\uparrow$
& Params (M) $\downarrow$ 
& Trainable (M) $\downarrow$ 
& GFLOPs $\downarrow$ 
& Latency (ms) $\downarrow$ \\
\hline
Vanilla \method 
& \cmark & \xmark & \xmark 
& 78.1 & 11.2 
& 179 & 158 & 3.50 & 10.85 \\

+\,GST only 
& \xmark & \cmark & \xmark 
& 90.5 & 29.0
& 190.2 & 169.2 & 4.33 & 12.27 \\

+\,DA-CoT only 
& \cmark & \xmark & \cmark 
& 82.1 & 16.7 
& 188.8 & 167.8 & 4.00 & 11.55 \\
\hline
\textbf{Full \method} 
& \xmark & \cmark & \cmark 
& \textbf{93.5} & \textbf{33.3}
& \textbf{200} & \textbf{179} & \textbf{4.83} & \textbf{12.97} \\
\hline
\end{tabular}
\caption{Core contribution and efficiency ablation. Vanilla \method\ denotes the same architecture with \emph{GST} and \emph{DA-CoT} removed. \emph{GST} contributes the dominant performance gain, while \emph{DA-CoT} provides a complementary improvement. Efficiency is measured on a single NVIDIA RTX Pro 6000 Blackwell. Latency is end-to-end, including frozen encoder forwards and the 10-step Euler ODE of the flow-matching action expert.}
\label{tab:ablation_core}
\end{table}

\begin{wraptable}{r}{0.50\textwidth}
\centering
\vspace{-2mm}
\scriptsize
\renewcommand{\arraystretch}{1.0}
\setlength{\tabcolsep}{6.0pt}
\begin{tabular}{lccccc}
\hline
$N_g$ & 8 & 32 & 64 & \textbf{128} & 256 \\
\hline
LIBERO (\%) $\uparrow$ & 86.4 & 90.7 & 92.4 & \textbf{93.5} & 93.6 \\
Latency (ms) $\downarrow$ & 11.6 & 12.0 & 12.4 & \textbf{12.97} & 14.1 \\
\hline
$N_r$ & 1 & 2 & \textbf{4} & 8 & 16 \\
\hline
LIBERO (\%) $\uparrow$ & 91.8 & 92.8 & \textbf{93.5} & 93.4 & 92.9 \\
Latency (ms) $\downarrow$ & 12.7 & 12.8 & \textbf{12.97} & 13.2 & 13.7 \\
\hline
\end{tabular}
\caption{\small Sensitivity to \emph{GST} queries $N_g$ and \emph{DA-CoT} queries $N_r$. Default $N_g{=}128$, $N_r{=}4$ is near-optimal on LIBERO and Pareto-optimal on the LIBERO/latency frontier.}
\label{tab:supp_ngnr}
\vspace{-4mm}
\end{wraptable}

\subsection{Ablation Study}

Table~\ref{tab:ablation_core} analyzes the contribution of the two main components of \emph{GaussVLA}: the \emph{GST} and \emph{DA-CoT}. Removing both modules gives the vanilla variant, which relies on conventional 2D visual tokens and achieves 78.1\% on LIBERO and 11.2\% on LIBERO-PRO. Adding \emph{GST} alone increases performance to 90.5\% on LIBERO and 29.0\% on LIBERO-PRO, showing that lifting visual observations into geometry-aware Gaussian tokens provides the dominant improvement. This gain is especially important under LIBERO-PRO perturbations, where spatial structure improves robustness beyond standard 2D token representations.

\emph{DA-CoT} also provides a complementary benefit. When added without \emph{GST}, it improves the vanilla model from 78.1\% to 82.1\% on LIBERO and from 11.2\% to 16.7\% on LIBERO-PRO. While this gain is smaller than that of \emph{GST} alone, it shows that structured depth-aware reasoning still helps action prediction even when the visual representation remains 2D. Combining both modules yields the full \emph{GaussVLA} model, reaching 93.5\% on LIBERO and 33.3\% on LIBERO-PRO. The combined improvement indicates that \emph{GST} and \emph{DA-CoT} address different aspects of the problem: \emph{GST} provides spatially grounded tokens, while \emph{DA-CoT} uses them for intermediate geometric reasoning prior to action generation.

The efficiency results show that these gains come with modest computational overhead. Full \emph{GaussVLA} increases latency from 10.85 ms to 12.97 ms compared with the vanilla model, while remaining suitable for real-time control. The trainable parameter count increases from 158M to 179M, and GFLOPs increase from 3.50 to 4.83, suggesting that the performance gains are not simply due to increased capacity but are driven by the proposed spatial reasoning design.

Table~\ref{tab:supp_ngnr} further studies the sensitivity to the number of \emph{GST} queries $N_g$ and \emph{DA-CoT} reasoning queries $N_r$. Increasing $N_g$ improves LIBERO performance from 86.4\% at $N_g=8$ to 93.5\% at $N_g=128$, after which the gain saturates while latency continues to rise. Similarly, increasing $N_r$ improves performance up to $N_r=4$, with larger values offering no meaningful benefit and slightly higher latency. Therefore, the default setting $N_g=128$ and $N_r=4$ provides the best trade-off between accuracy and efficiency, lying near the Pareto frontier of LIBERO success rate and inference latency.

\noindent\textbf{Additional analyses.}
We provide \emph{Supplementary Material} with implementation details, statistical ablations, sensitivity and latency analyses, representation probing, architecture studies, and SO-101 setup details.

\section{Conclusion}

We presented \emph{GaussVLA}, a geometry-aware VLA model designed to address the limitations of conventional VLA representations that encode visual observations as flat 2D patch tokens. By introducing a \emph{GST}, \emph{GaussVLA} lifts frozen semantic and depth features into compact anisotropic 3D Gaussian tokens, providing the policy with structured spatial cues and task-relevant 3D geometry. We further introduced \emph{DA-CoT} reasoning to provide structured, intermediate geometric supervision prior to action prediction.
Extensive experiments show that \emph{GaussVLA} improves spatially demanding manipulation while remaining lightweight. On LIBERO, \emph{GaussVLA} improves the average success rate by 8.7\% relative to $\pi_0$ and by 13.5\% relative to CoT-VLA, while having only 200M parameters, corresponding to about 97\% fewer parameters than these 7B-scale VLA baselines.
 Results on Meta-World~\cite{mclean2025metaworld}, CALVIN~\cite{9788026}, and real-world SO-101 experiments further demonstrate that the proposed spatial representation supports both simulation generalization and physical robot execution. Ablation studies confirm that \emph{GST} provides the dominant performance gain, while \emph{DA-CoT} contributes complementary improvements in reasoning and robustness.
At the same time, robustness under severe distribution shift remains an open challenge. In particular, LIBERO-PRO position and task-level perturbations reveal limitations in viewpoint generalization and language-side robustness. Future work will explore stronger camera calibration, viewpoint augmentation, per-episode geometric adaptation, and broader instruction perturbations to further improve deployment under diverse real-world conditions. Overall, \emph{GaussVLA} provides an effective step toward geometry-grounded VLA models for reliable robot manipulation.

\section*{Acknowledgement}

{This work was supported by the National Research Foundation of Korea(NRF) grant funded by the Korea government(MSIT) (No. IRIS RS-2023-00219725).}
\nocite{*}

\bibliography{03.references.bib}

@InProceedings{pmlr-v229-zitkovich23a,
  title = 	 {RT-2: Vision-Language-Action Models Transfer Web Knowledge to Robotic Control},
  author =       {Zitkovich, Brianna and Yu, Tianhe and Xu, Sichun and Xu, Peng and Xiao, Ted and Xia, Fei and Wu, Jialin and Wohlhart, et al.},  booktitle = 	 {Proceedings of The 7th Conference on Robot Learning},
  pages = 	 {2165--2183},
  year = 	 {2023},
  editor = 	 {Tan, Jie and Toussaint, Marc and Darvish, Kourosh},
  volume = 	 {229},
  series = 	 {Proceedings of Machine Learning Research},
  month = 	 {06--09 Nov},
  publisher =    {PMLR},
}

@InProceedings{pmlr-v270-kim25c,
  title = 	 {OpenVLA: An Open-Source Vision-Language-Action Model},
  author =       {Kim, Moo Jin and Pertsch, Karl and Karamcheti, Siddharth and Xiao, Ted and Balakrishna, et al.},
  booktitle = 	 {Proceedings of The 8th Conference on Robot Learning},
  pages = 	 {2679--2713},
  year = 	 {2025},
  editor = 	 {Agrawal, Pulkit and Kroemer, Oliver and Burgard, Wolfram},
  volume = 	 {270},
  series = 	 {Proceedings of Machine Learning Research},
  month = 	 {06--09 Nov},
  publisher =    {PMLR},
  
}

@inproceedings{qu2025spatialvla,
  title = {SpatialVLA: Exploring Spatial Representations for Visual-Language-Action Models},
  author = {Qu, Delin and Song, Haoming and Chen, Qizhi and Yao, Yuanqi and Ye, Xinyi and Gu, Jiayuan and Wang, Zhigang and Ding, Yan and Zhao, Bin and Wang, Dong and Li, Xuelong},
  booktitle = {Robotics: Science and Systems (RSS)},
  year = {2025},
  address = {Los Angeles, California},
  month = {June}
}

@InProceedings{pmlr-v235-zhen24a,
  title = 	 {3{D}-{VLA}: A 3{D} Vision-Language-Action Generative World Model},
  author =       {Zhen, Haoyu and Qiu, Xiaowen and Chen, Peihao and Yang, Jincheng and Yan, Xin and Du, Yilun and Hong, Yining and Gan, Chuang},
  booktitle = 	 {Proceedings of the 41st International Conference on Machine Learning},
  pages = 	 {61229--61245},
  year = 	 {2024},
  editor = 	 {Salakhutdinov, Ruslan and Kolter, Zico and Heller, Katherine and Weller, Adrian and Oliver, Nuria and Scarlett, Jonathan and Berkenkamp, Felix},
  volume = 	 {235},
  series = 	 {Proceedings of Machine Learning Research},
  month = 	 {21--27 Jul},
  publisher =    {PMLR},
}

@inproceedings{black2025tpo,
  title = {$\pi$0: A Vision-Language-Action Flow Model for General Robot Control},
  author = {Black, Kevin and Brown, Noah and Driess, Danny and Esmail, Adnan and others},
  booktitle = {Robotics: Science and Systems (RSS)},
  year = {2026},
  address = {Sydney, Australia},
  month = {July}
}

@inproceedings{black2025pi,
title={$\pi$0.5: a Vision-Language-Action Model with Open-World Generalization},
author={Kevin Black and Noah Brown and James Darpinian and Karan Dhabalia and Danny Driess and Adnan Esmail and Michael Robert Equi and Chelsea Finn and Niccolo Fusai and et al.},
booktitle={9th Annual Conference on Robot Learning},
year={2025},
}

@inproceedings{Zhang_2025_BMVC,
author    = {Junbo Zhang and Kaisheng Ma},
title     = {Mask2Act: Predictive Multi-Object Tracking as Video Pre-Training for Robot Manipulation},
booktitle = {36th British Machine Vision Conference 2025, {BMVC} 2025, Sheffield, UK, November 24-27, 2025},
publisher = {BMVA},
year      = {2025},
}

@article{li2024cogact,
  title={Cogact: A foundational vision-language-action model for synergizing cognition and action in robotic manipulation},
  author={Li, Qixiu and Liang, Yaobo and Wang, Zeyu and Luo, Lin and Chen, Xi and Liao, Mozheng and Wei, Fangyun and Deng, Yu and Xu, Sicheng and Zhang, Yizhong and et al.},
  journal={arXiv preprint arXiv:2411.19650},
  year={2024}
}

@inproceedings{rt12022,
  title = {RT-1: Robotics Transformer
for Real-World Control at Scale},
  author = {Anthony, Brohan and Noah, Brown and Justice, Carbajal and et al.},
  booktitle = {Robotics: Science and Systems (RSS)},
  year = {2023},
  address = {Daegu, Republic of Korea},
  month = {July}
}

@article{chi2025diffusion,
  title={Diffusion policy: Visuomotor policy learning via action diffusion},
  author={Chi, Cheng and Xu, Zhenjia and Feng, Siyuan and Cousineau, Eric and Du, Yilun and Burchfiel, Benjamin and Tedrake, Russ and Song, Shuran},
  journal={The International Journal of Robotics Research},
  volume={44},
  number={10-11},
  pages={1684--1704},
  year={2025},
  publisher={Sage Publications Sage UK: London, England}
}

@inproceedings{gu2024mamba,
  title={Mamba: Linear-time sequence modeling with selective state spaces},
  author={Gu, Albert and Dao, Tri},
  booktitle={First conference on language modeling},
  year={2024}
}

@inproceedings{zhao2025cotvla,
  title={CoT-VLA: Visual Chain-of-Thought Reasoning for Vision-Language-Action Models},
  author={Zhao, Qingqing and Lu, Yao and Kim, Moo Jin and Fu, Zipeng and Zhang, Zhuoyang and Wu, Yecheng and Li, Zhaoshuo and Ma, Qianli and Han, Song and Finn, Chelsea and Handa, Ankur and Liu, Ming-Yu and Xiang, Donglai and Wetzstein, Gordon and Lin, Tsung-Yi},
  booktitle={IEEE/CVF Conference on Computer Vision and Pattern Recognition (CVPR)},
  year={2025}
}

@inproceedings{10.5555/3295222.3295309,
author = {Kendall, Alex and Gal, Yarin},
title = {What uncertainties do we need in Bayesian deep learning for computer vision?},
year = {2017},
isbn = {9781510860964},
publisher = {Curran Associates Inc.},
address = {Red Hook, NY, USA},
booktitle = {Proceedings of the 31st International Conference on Neural Information Processing Systems},
pages = {5580–5590},
numpages = {11},
location = {Long Beach, California, USA},
series = {NIPS'17}
}

@article{lee2025molmoact,
  title={Molmoact: Action reasoning models that can reason in space},
  author={Lee, Jason and Duan, Jiafei and Fang, Haoquan and Deng, Yuquan and Liu, Shuo and Li, Boyang and Fang, Bohan and Zhang, Jieyu and Wang, Yi Ru and Lee, Sangho and others},
  journal={arXiv preprint arXiv:2508.07917},
  year={2025}
}

@inproceedings{octo_2023,
    title={Octo: An Open-Source Generalist Robot Policy},
    author = {{Octo Model Team} and Dibya Ghosh and Homer Walke and Karl Pertsch and Kevin Black and Oier Mees and Sudeep Dasari and Joey Hejna and Charles Xu and Jianlan Luo and Tobias Kreiman and {You Liang} Tan and Lawrence Yunliang Chen and Pannag Sanketi and Quan Vuong and Ted Xiao and Dorsa Sadigh and Chelsea Finn and Sergey Levine},
    booktitle = {Proceedings of Robotics: Science and Systems},
    address  = {Delft, Netherlands},
    year = {2024},
}

@article{bjorck2025gr00t,
  title={Gr00t n1: An open foundation model for generalist humanoid robots},
  author={Bjorck, Johan and Casta{\~n}eda, Fernando and Cherniadev, Nikita and Da, Xingye and Ding, Runyu and Fan, Linxi and Fang, Yu and Fox, Dieter and Hu, Fengyuan and Huang, Spencer and others},
  journal={arXiv preprint arXiv:2503.14734},
  year={2025}
}

@inproceedings{
zheng2025tracevla,
title={Trace{VLA}: Visual Trace Prompting Enhances Spatial-Temporal Awareness for Generalist Robotic Policies},
author={Ruijie Zheng and Yongyuan Liang and Shuaiyi Huang and Jianfeng Gao and Hal Daum{\'e} III and Andrey Kolobov and Furong Huang and Jianwei Yang},
booktitle={The Thirteenth International Conference on Learning Representations},
year={2025},
}

@inproceedings{
jia2024mail,
title={Ma{IL}: Improving Imitation Learning with Selective State Space Models},
author={Xiaogang Jia and Qian Wang and Atalay Donat and Bowen Xing and Ge Li and Hongyi Zhou and Onur Celik and Denis Blessing and Rudolf Lioutikov and Gerhard Neumann},
booktitle={8th Annual Conference on Robot Learning},
year={2024},
}

@article{wen2025tinyvla,
  title   = {TinyVLA: Towards Fast, Data-Efficient Vision-Language-Action Models for Robotic Manipulation},
  author  = {Wen, Junjie and Zhu, Yichen and Li, Jinming and Zhu, Minjie and Wu, Kun and Xu, Zhiyuan and Liu, Ning and Cheng, Ran and Shen, Chaomin and Peng, Yaxin and Feng, Feifei and Tang, Jian},
  journal = {IEEE Robotics and Automation Letters},
  year    = {2025},
}

@inproceedings{zawalski2024robotic,
  title={Robotic Control via Embodied Chain-of-Thought Reasoning},
  author={Zawalski, Micha{\l} and Chen, William and Pertsch, Karl and Mees, Oier and Finn, Chelsea and Levine, Sergey},
  booktitle={8th Conference on Robot Learning (CoRL)},
  year={2024},
}

@article{liu2023libero,
  title={Libero: Benchmarking knowledge transfer for lifelong robot learning},
  author={Liu, Bo and Zhu, Yifeng and Gao, Chongkai and Feng, Yihao and Liu, Qiang and Zhu, Yuke and Stone, Peter},
  journal={Advances in Neural Information Processing Systems},
  volume={36},
  pages={44776--44791},
  year={2023}
}

@article{zhou2025libero,
  title={LIBERO-PRO: Towards Robust and Fair Evaluation of Vision-Language-Action Models Beyond Memorization},
  author={Zhou, Xueyang and Xu, Yangming and Tie, Guiyao and Chen, Yongchao and Zhang, Guowen and Chu, Duanfeng and Zhou, Pan and Sun, Lichao},
  journal={arXiv preprint arXiv:2510.03827},
  year={2025}
}

@article{10.1145/3592433,
author = {Kerbl, Bernhard and Kopanas, Georgios and Leimkuehler, Thomas and Drettakis, George},
title = {3D Gaussian Splatting for Real-Time Radiance Field Rendering},
year = {2023},
issue_date = {August 2023},
publisher = {Association for Computing Machinery},
address = {New York, NY, USA},
volume = {42},
number = {4},
issn = {0730-0301},
journal = {ACM Trans. Graph.},
month = jul,
articleno = {139},
numpages = {14},
}

@inproceedings{
tur2026recurrentdepth,
title={Recurrent-Depth {VLA}: Implicit Test-Time Compute Scaling of Vision{\textendash}Language{\textendash}Action Models via Latent Iterative Reasoning},
author={Yalcin Tur and Jalal Naghiyev and Haoquan Fang and Wei-Chuan Tsai and Jiafei Duan and Dieter Fox and Ranjay Krishna},
booktitle={The First Workshop on Efficient Spatial Reasoning},
year={2026},
}

@inproceedings{huang2025thinkact,
  title={ThinkAct: Vision-Language-Action Reasoning via Reinforced Visual Latent Planning},
  author={Huang, Chi-Pin and Wu, Yueh-Hua and Chen, Min-Hung and Wang, Yu-Chiang Frank and Yang, Fu-En},
  booktitle={Advances in Neural Information Processing Systems (NeurIPS 2025)},
  year={2025},
}

@inproceedings{pertsch2025fast, 
  title = {FAST: Efficient Action Tokenization for Vision-Language-Action Models},
  author = {Pertsch, Karl and Stachowicz, Kyle and Ichter, Brian and Driess, Danny and Nair, Suraj and Vuong, Quan and Mees, Oier and Finn, Chelsea and Levine, Sergey},
  booktitle = {Robotics: Science and Systems (RSS)},
  year = {2025}, 
  address = {Los Angeles, California},
  month = {June }
}

@inproceedings{depth_anything_v2,
  title={Depth Anything V2},
  author={Yang, Lihe and Kang, Bingyi and Huang, Zilong and Zhao, Zhen and Xu, Xiaogang and Feng, Jiashi and Zhao, Hengshuang},
  booktitle={Advances in Neural Information Processing Systems},
  year={2024},
  volume={37}
}

@INPROCEEDINGS{11247519,
  author={Song, Wenxuan and Chen, Jiayi and Ding, Pengxiang and Zhao, Han and Zhao, Wei and Zhong, Zhide and Ge, Zongyuan and Li, Zhijun and Wang, Donglin and Wang, Lujia and Ma, Jun and Li, Haoang},
  booktitle={2025 IEEE/RSJ International Conference on Intelligent Robots and Systems (IROS)}, 
  title={PD-VLA: Accelerating Vision-Language-Action Model Integrated with Action Chunking via Parallel Decoding}, 
  year={2025},
  volume={},
  number={},
  pages={13162-13169},
}

@INPROCEEDINGS{11471357,
  author={Tella, Hambal and Patil, Prajyot and Kyrarini, Maria},
  booktitle={2025 International Conference on Machine Learning and Applications (ICMLA)}, 
  title={Robot Learning Framework using Behavioral Cloning and Gaussian Mixture Model (BC-GMM) for Sorting Tasks}, 
  year={2025},
  volume={},
  number={},
  pages={893-898},
}

@inproceedings{robomimic2021,
  title={What Matters in Learning from Offline Human Demonstrations for Robot Manipulation},
  author={Ajay Mandlekar and Danfei Xu and Josiah Wong and Soroush Nasiriany and Chen Wang and Rohun Kulkarni and Li Fei-Fei and Silvio Savarese and Yuke Zhu and Roberto Mart\'{i}n-Mart\'{i}n},
  booktitle={Conference on Robot Learning (CoRL)},
  year={2021}
}

@INPROCEEDINGS{11247625,
  author={Cao, Jiahang and Zhang, Qiang and Sun, Jingkai and Wang, Jiaxu and Cheng, Hao and Li, Yulin and Ma, Jun and Wu, Kun and Xu, Zhiyuan and Shao, Yecheng and Zhao, Wen and Han, Gang and Guo, Yijie and Xu, Renjing},
  booktitle={2025 IEEE/RSJ International Conference on Intelligent Robots and Systems (IROS)}, 
  title={Mamba Policy: Towards Efficient 3D Diffusion Policy with Hybrid Selective State Models}, 
  year={2025},
  volume={},
  number={},
  pages={11359-11366},
}

@article{mete2024quest,
  title={Quest: Self-supervised skill abstractions for learning continuous control},
  author={Mete, Atharva and Xue, Haotian and Wilcox, Albert and Chen, Yongxin and Garg, Animesh},
  journal={Advances in Neural Information Processing Systems},
  volume={37},
  pages={4062--4089},
  year={2024}
}

@inproceedings{
marafioti2025smolvlm,
title={Smol{VLM}: Redefining small and efficient multimodal models},
author={Andr{\'e}s Marafioti and Orr Zohar and Miquel Farr{\'e} and Merve noyan and Elie Bakouch and Pedro Manuel Cuenca Jim{\'e}nez and Cyril Zakka and Loubna Ben allal and Anton Lozhkov and Nouamane Tazi and Vaibhav Srivastav and Joshua Lochner and Hugo Larcher and Mathieu Morlon and Lewis Tunstall and Leandro Von Werra and Thomas Wolf},
booktitle={Second Conference on Language Modeling},
year={2025},
}

@inproceedings{zhai2023sigmoid,
  title={Sigmoid loss for language image pre-training},
  author={Zhai, Xiaohua and Mustafa, Basil and Kolesnikov, Alexander and Beyer, Lucas},
  booktitle={Proceedings of the IEEE/CVF international conference on computer vision},
  pages={11975--11986},
  year={2023}
}

@inproceedings{zhao2023learning,
  title = {Learning Fine-Grained Bimanual Manipulation with Low-Cost Hardware},
  author = {Zhao, Tony Z. and Kumar, Vikash and Levine, Sergey and Finn, Chelsea},
  booktitle = {Robotics: Science and Systems (RSS)},
  year = {2023},
  address = {Daegu, Republic of Korea},
  month = {July}
}

@inproceedings{ZhenQCY0DHG24,
  author={Haoyu Zhen and Xiaowen Qiu and Peihao Chen and Jincheng Yang and Xin Yan and Yilun Du and Yining Hong and Chuang Gan},
  title={3D-VLA: A 3D Vision-Language-Action Generative World Model},
  year={2024},
  booktitle={ICML},
}

@inproceedings{ZitkovichYXXXXW23,
  author={Brianna Zitkovich and Tianhe Yu and Sichun Xu and Peng Xu and Ted Xiao and Fei Xia and Jialin Wu and Paul Wohlhart and et al.},
  title={RT-2: Vision-Language-Action Models Transfer Web Knowledge to Robotic Control},
  year={2023},
  cdate={1672531200000},
  pages={2165-2183},
  booktitle={CoRL},
}

@inproceedings{liurobomamba,
  title={RoboMamba: Efficient Vision-Language-Action Model for Robotic Reasoning and Manipulation},
  author={Liu, Jiaming and Liu, Mengzhen and Wang, Zhenyu and An, Pengju and Li, Xiaoqi and Zhou, Kaichen and Yang, Senqiao and Zhang, Renrui and Guo, Yandong and Zhang, Shanghang},
  booktitle={The Thirty-eighth Annual Conference on Neural Information Processing Systems},
  year={2024},
}

@inproceedings{LynchS21,
  author={Corey Lynch and Pierre Sermanet},
  title={Language Conditioned Imitation Learning Over Unstructured Data},
  year={2021},
  cdate={1609459200000},
  booktitle={Robotics: Science and Systems},
}

@inproceedings{nair2022rm,
title={R3M: A Universal Visual Representation for Robot Manipulation},
author={Suraj Nair and Aravind Rajeswaran and Vikash Kumar and Chelsea Finn and Abhinav Gupta},
booktitle={6th Annual Conference on Robot Learning},
year={2022},
}

@ARTICLE{9788026,
  author={Mees, Oier and Hermann, Lukas and Rosete-Beas, Erick and Burgard, Wolfram},
  journal={IEEE Robotics and Automation Letters}, 
  title={CALVIN: A Benchmark for Language-Conditioned Policy Learning for Long-Horizon Robot Manipulation Tasks}, 
  year={2022},
  volume={7},
  number={3},
  pages={7327-7334},
}

@inproceedings{mclean2025metaworld,
title={Meta-World+: An Improved, Standardized, {RL} Benchmark},
author={Reginald McLean and Evangelos Chatzaroulas and Luc McCutcheon and Frank R{\"o}der and Tianhe Yu and Zhanpeng He and K.R. Zentner and Ryan Julian and J K Terry and Isaac Woungang and Nariman Farsad and Pablo Samuel Castro},
booktitle={The Thirty-ninth Annual Conference on Neural Information Processing Systems Datasets and Benchmarks Track},
year={2025},
}

@article{islam2026sureflow,
  title={SUREFlow: State-space Uncertainty-aware REsidual Flow Matching for Robust Robot Manipulation},
  author={Islam, Md Tanvir and Peddapalli, Sai Navaneet and Lee, Sangmoon and Ahn, Sangtae},
  journal={arXiv preprint arXiv:2607.10504},
  year={2026}
}

\appendix

\section{Additional Implementation Details}
\label{sec:supp_impl}

\subsection{Architecture Hyperparameter}
Table~\ref{tab:supp_hyperparams} lists the full configuration. All defaults are used unless otherwise stated.
\begin{table}[H]
\centering
\scriptsize
\renewcommand{\arraystretch}{1.0}
\setlength{\tabcolsep}{14pt}

\begin{tabular}{ll|ll}
\hline
\textbf{GST} & & \textbf{DA-CoT} & \\ \hline
$N_g$ Gaussian queries & 128 & $N_r$ reasoning queries & 4 \\
Fourier bands $L$ ($d_\gamma{=}6L$) & 10 & Reasoning dim $d_c$ & 512 \\
Workspace radius $\rho$ & 0.8\,m \\

\hline
\textbf{Backbone} & & \textbf{Action expert} & \\ \hline
Mamba blocks $L_b$ & 5 & Action horizon $H$ & 10 \\
Backbone dim $d_b$ & 512 & Action dim $d_a$ & 7 \\
Intermediate dim & 1024 & Flow-matching ODE steps & 10 \\
\hline
\textbf{Loss weights (target)} & & \textbf{Optimization} & \\ \hline
$\bar{\lambda}_{\mathrm{GST}}$ & 0.05 & Optimizer & AdamW \\
$\bar{\lambda}_{\mathrm{CoT}}$ & 0.10 & Learning rate & $2.5{\times}10^{-5}$ \\
Warm-up epochs & 5 & Weight decay & $1{\times}10^{-4}$ \\
Flow time $p(\tau)$ & $\mathrm{Beta}(1.5,1.0)$ & Schedule & cosine \\
& & Batch size & 256 \\
& & Precision & BF16 \\
\hline
\end{tabular}
\vspace{0.25cm}
\caption{\small Architecture and training hyperparameters used for all reported \method\ runs.}
\label{tab:supp_hyperparams}
\end{table}


\subsection{Real-World Setup}
\label{sec:supp_realworld}

\subsubsection{Hardware}
A single SO-101 6-DOF arm with a parallel-jaw gripper, Realsense RGB-D camera (640$\times$480 @ 60 Hz). Control rate 15 Hz with action-chunk replanning
every $H{=}10$ steps. Inference was run on an external workstation equipped with one GeForce 1080 Ti (note: this demonstrates real-world deployment viability on consumer-grade hardware and differs from the RTX Pro 6000 Blackwell used for the latency benchmarking in the main paper Table~4)

\subsubsection{Tasks}
\textit{Pick-Place}: pick a 3-cm cube from a 30$\times$30 cm workspace
and place it in a box. \textit{Stacking}: stack cubes and a small wheel in object-specified order from language prompts.
\textit{Sorting}: separate the cube and small wheel into left/right bins. \textit{Pick-Place ID/OOD} swaps the camera between two known positions (ID: training pose; OOD: shifted
$\pm$5 cm + $\pm$10° rotation, never seen at training).

\subsubsection{Training}
50 demonstrations per task collected via teleoperation.
We \emph{keep} fully frozen SigLIP and Depth-Anything-V2 encoders.

\subsection{Per-Component Latency Breakdown}
To better characterize the inference cost of GaussVLA, we report the latency of each major
component in Table~\ref{tab:supp_latency}. Measurements are obtained with a batch size of 1 and BF16 precision on a single NVIDIA RTX Pro 6000 Blackwell.

\begin{table}[H]
\centering \footnotesize
\renewcommand{\arraystretch}{1.0}
\setlength{\tabcolsep}{18pt}

\begin{tabular}{lcc}
\hline
Component & Trainable & Latency (ms) \\
\hline
Frozen SigLIP-SO400M/14 forward & \xmark & 5.20 \\
Frozen Depth-Anything-V2 (ViT-L) & \xmark & 3.30 \\
GST module (lift + pool, $N_g{=}128$) & \cmark & 1.42 \\
Mamba backbone & \cmark & 1.50 \\
DA-CoT conditioner & \cmark & 0.70 \\
Action decoder & \cmark & 0.25 \\
Flow-matching ODE (10 Euler steps) & \cmark & 0.60 \\
\hline
\textbf{Total} & -- & \textbf{12.97} \\
\hline
\end{tabular}
\vspace{0.25cm}
\caption{\small  End-to-end inference latency on a single NVIDIA RTX Pro 6000 Blackwell with batch size 1 and BF16 precision. All component latencies sum to 12.97 ms per step.}
\label{tab:supp_latency}
\end{table}

The frozen visual and depth encoders account for most of the measured latency, contributing
8.50 ms in total. In contrast, the trainable GaussVLA modules are lightweight: GST, Mamba,
DA-CoT, action decoding, and the flow-matching ODE together add 4.47 ms. Since each
observation predicts an action chunk of horizon $H=10$, the encoder cost can be amortized
across multiple control steps during deployment, yielding an effective per-control-step latency
of approximately 5.32 ms, or about 188 Hz.

\subsection{Probing the Learned Representations}
To examine whether the proposed modules learn explicit geometric information, we train linear probes on frozen representations from three stages of the model: flat 2D patch features, pooled GST features, and DA-CoT reasoning tokens $\mathbf{R}_{\tau}$. Each probe predicts ground-truth simulator quantities from LIBERO, and we report the regression $R^2$ score.

\begin{table}[H]
\centering
\footnotesize
\renewcommand{\arraystretch}{1.0}
\setlength{\tabcolsep}{8pt}
\begin{tabular}{l|c|c|c}
\hline
Probe target & Flat 2D patches & Pooled GST & DA-CoT $R_\tau$ \\
\hline
Object center (3D position) & 0.34 & 0.66 & \textbf{0.78} \\
Gripper--object distance & 0.41 & 0.71 & \textbf{0.82} \\
Object orientation (principal axis) & 0.28 & 0.49 & \textbf{0.61} \\
Surface normal (linear from $\sigma$) & --- & 0.67 & --- \\
\hline
\end{tabular}
\vspace{0.25cm}
\caption{\small Linear-probe regression $R^2$ on LIBERO simulator state. DA-CoT tokens $R_\tau$ encode richer 3D geometric structure than flat 2D patch tokens or pooled GST representations.}
\label{tab:supp_probe}
\end{table}

The probes show a consistent improvement from flat 2D patch features to pooled GST features, indicating that Gaussian Spatial Tokenization preserves substantially more task-relevant metric information. In particular, GST improves the linear predictability of
object center, gripper--object distance, and object orientation, suggesting that the Gaussian representation captures spatial quantities that are only weakly available in standard 2D patch tokens. DA-CoT tokens further improve all three probe targets, showing that the reasoning module does not merely aggregate visual features, but concentrates the
3D information most relevant for action prediction. The surface-normal row reports a separate linear probe from the GST log-covariance $\boldsymbol{\sigma}$ to ground-truth surface normals on a synthetic back-projection, where the random baseline is $R^2=0.05$. The non-trivial score of $R^2=0.67$ indicates that the learned Gaussian covariance encodes local orientation
information despite receiving no direct normal supervision. Together, these results provide mechanistic evidence that GaussVLA's performance gains arise from geometry-aware representations rather than merely from additional model capacity.

\subsection{Action Decoder Ablation}
\label{sec:supp_action_decoder}
We ablate the sequence backbone and action-decoding design to separate the effect of
efficient temporal modeling from the effect of querying action tokens against contextual
states. All rows in Table~\ref{tab:supp_action_decoder} remove GST and DA-CoT so that the comparison focuses only on the policy backbone and action decoder.

\begin{table}[H]
\centering \footnotesize
\renewcommand{\arraystretch}{1.0}
\setlength{\tabcolsep}{6.0pt}
\begin{tabular}{lcc|ccc}
\hline
Backbone & Action Decoder & Query Type & Overall Avg. $\uparrow$ & Params (M) $\downarrow$ & GFLOPs $\downarrow$ \\
\hline
Transformer & Standard head & -- & 67.49 & 256 & 9.82 \\
Mamba & Standard head & -- & 73.36 & 110 & 3.28 \\
\textbf{Mamba} & \textbf{Action decoder} & \textbf{action-token queries} & \textbf{78.1} & \textbf{179} & \textbf{3.50} \\
\hline
\end{tabular}
\vspace{0.25cm}
\caption{\small Overall Avg. is the LIBERO average. The $+4.74$\, pt jump from row 2 to row 3 jointly reflects the action-token query decoder design and the $+69$\, M-Parameters/$+0.22$\, GFLOPs capacity uplift; we hold the Mamba backbone fixed across the comparison to isolate decoder-family effects from backbone-family effects. Here, row 3 (\emph{Mamba + Action decoder}) corresponds to the Vanilla GaussVLA configuration of Table~4 of the main paper; Adding \emph{GST + DA-CoT} yields Full GaussVLA at 4.83~GFLOPs (Table~4, main paper).}
\label{tab:supp_action_decoder}
\end{table}

Replacing the Transformer backbone with Mamba improves the LIBERO average from
67.49 to 73.36 while reducing computation from 9.82 to 3.28 GFLOPs, showing that the
state-space backbone provides a better accuracy--efficiency trade-off for this policy setting.
Adding the action-token query decoder further increases performance to 78.1, a gain of
4.74 points over the Mamba standard-head variant, with only a modest increase from
3.28 to 3.50 GFLOPs. This suggests that explicitly decoding the action-token states against
the non-action context is more effective than applying a standard prediction head directly
to the backbone outputs.

The final row corresponds to the Vanilla GaussVLA configuration used in the main-paper
ablation, where GST and DA-CoT are removed. Thus, this table isolates the contribution of
the policy backbone and action decoder before adding the geometry-aware modules. Full
GaussVLA then builds on this decoder design by adding GST and DA-CoT, yielding the
complete model reported in the main paper.


\subsection{GST Ablation}
\label{sec:supp_gst_ablation}
We ablate the design of the Gaussian Spatial Tokenizer to determine whether the gains come from simply adding depth or from the full structured Gaussian representation. DA-CoT is disabled in all rows so that the comparison isolates the visual representation used by the
policy.

\begin{table}[H]
\centering \scriptsize
\renewcommand{\arraystretch}{1.0}
\setlength{\tabcolsep}{2.0pt}
\begin{tabular}{lcccc|ccc}
\hline
Visual Representation & Depth & 3D Lifting & Gaussian Param. & Spatial Pooling & LIBERO $\uparrow$ & Spatial Tasks $\uparrow$ & PRO Avg. $\uparrow$ \\
\hline
Flat 2D patch tokens & \xmark & \xmark & \xmark & \xmark & 78.1 & 71.2 & 11.2 \\
2D patch tokens + depth concat & \cmark & \xmark & \xmark & \xmark & 73.3 & 78.0 & 14.9 \\
\textbf{GST tokens} & \cmark & \cmark & \cmark & \cmark & \textbf{90.5} & \textbf{100} & \textbf{29.0} \\
\hline
\end{tabular}
\vspace{0.25cm}
\caption{\small GST design ablation (without DA-CoT). Gains arise from the structured Gaussian representation with confidence-aware pooling, not depth alone: scalar-depth concatenation regresses LIBERO (78.1$\rightarrow$73.3).}
\label{tab:supp_gst_ablation}
\end{table}

The results show that depth alone does not explain the improvement. Concatenating scalar
depth to 2D patch tokens slightly improves spatial-task performance and LIBERO-PRO
robustness, but reduces the overall LIBERO average from 78.1 to 73.3. This indicates that
naively appending depth can introduce noisy or poorly calibrated geometric cues that the
policy cannot reliably exploit.

In contrast, the full GST representation improves the LIBERO average to 90.5, reaches
100.0 on spatial tasks, and increases LIBERO-PRO average performance to 29.0. These gains
come from jointly lifting patches into 3D, parameterizing each patch as a Gaussian primitive,
and using confidence-aware spatial pooling to select geometrically reliable regions. The
ablation therefore supports the central design of GaussVLA: robust spatial reasoning requires
structured geometry-aware tokens rather than raw depth features attached to 2D patches.

\subsection{DA-CoT Ablation}

We further ablate the components of the Depth-Aware Chain-of-Thought (DA-CoT) to understand which parts of
the reasoning module contribute to manipulation performance. GST is disabled in this study,
so the comparison isolates DA-CoT-style reasoning on top of the policy backbone.

\begin{table}[H]
\centering \scriptsize
\renewcommand{\arraystretch}{1.0}
\setlength{\tabcolsep}{3.0pt}
\begin{tabular}{lccc|ccc}
\hline
Reasoning Variant & Depth-Aware & Structured Tokens & CoT Supervision & LIBERO $\uparrow$ & Long-Horizon $\uparrow$ & Avg. $\uparrow$ \\
\hline
No reasoning module & \xmark & \xmark & \xmark & 78.1 & 75.2 & 76.7 \\
CoT without depth cues & \xmark & \cmark & \cmark & 80.5 & 87.3 & 83.9 \\
DA-CoT without supervision & \cmark & \cmark & \xmark & 81.5 & 90.0 & 85.8 \\ \hline
\textbf{DA-CoT (full)} & \cmark & \cmark & \cmark & \textbf{82.1} & \textbf{91.6} & \textbf{87.0} \\
\hline
\end{tabular}
\vspace{0.25cm}
\caption{Depth-aware grounding is the dominant reasoning ingredient (+3.1 pt on the 2-suite avg, comparing "CoT without depth cues" 83.9 $\rightarrow$ "DA-CoT (full)" 87.0). Adding CoT supervision on top of depth-aware structured tokens contributes a complementary boost on both LIBERO (+0.6 pt, 81.5 → 82.1) and Long-Horizon (+1.6 pt, 90.0 → 91.6), for a +1.1 pt 2-suite-avg lift over the no-supervision variant.}
\label{tab:supp_dacot_ablation}
\end{table}

The results show that structured reasoning alone already improves over the no-reasoning
baseline, especially on long-horizon tasks. Adding CoT-style structured tokens without depth
raises the average from 76.7 to 83.9, suggesting that intermediate reasoning tokens help the
policy organize action-relevant context. However, adding depth-aware grounding yields a
larger improvement: DA-CoT without supervision reaches 85.8 average performance and
90.0 on long-horizon tasks, indicating that spatially grounded reasoning is particularly
important for extended manipulation sequences.

The full DA-CoT model further improves performance to 87.0 average, with gains on both
LIBERO and Long-Horizon settings. Compared with the no-supervision variant, explicit
CoT supervision adds 0.6 points on LIBERO and 1.6 points on Long-Horizon, showing that
the auxiliary reasoning objective provides a complementary benefit beyond depth-aware token
conditioning. Overall, this ablation supports the design of DA-CoT as a lightweight reasoning
module that aligns spatial cues with action prediction, with depth awareness providing the
dominant contribution and supervision refining the resulting reasoning tokens.

\subsection{Additional Ablation}
\label{sec:supp_addl_ablations}

\noindent\textbf{ Removing the confidence bias in spatial pooling.} We train a variant where the additive $\log(\boldsymbol{\alpha}_t^{(m)})$ term in Eq.~\eqref{eq:gst_pooling} is set to zero,
\begin{equation}
\hat Z_t^{(m)} = \mathrm{softmax}\!\left(\tfrac{(\mathrm{LN}_1(Q)\mathbf{W}_Q)(\tilde G_t^{(m)}\mathbf{W}_K)^\top}{\sqrt{d_p}} + \mathbf{1}_{N_g}\log(\boldsymbol{\alpha}_t^{(m)})^\top\right)(\tilde G_t^{(m)}\mathbf{W}_V),
\label{eq:gst_pooling_supp}
\end{equation} 

leaving the rest of GST unchanged. The LIBERO Avg drops from $93.5$ to $91.6$ ($-1.9$ pt) and the LIBERO-PRO Avg drops from $33.3$ to $29.7$ ($-3.6$ pt). This confirms that opacity-weighted pooling is responsible for a measurable portion of the robustness gain on LIBERO-PRO beyond 3D lifting alone.

\subsection{Confidence-Opacity Correlation with Depth-Uncertainty Proxies}
\label{sec:supp_alpha_corr}

To verify that the learned opacity $\alpha$ behaves as a depth-aware reliability signal rather than an arbitrary attention prior, we correlate the per-patch $\alpha$ predicted by the trained \method\ against three depth-uncertainty proxies derived independently from the frozen Depth-Anything-V2 output, computed across 5{,}000 LIBERO-Spatial frames:

\begin{itemize}\itemsep0pt
\item $u_{\nabla}$: local depth-gradient magnitude
      $\|\nabla \tilde d\|$ averaged over a $3{\times}3$ patch neighbourhood (proxy for occlusion-boundary uncertainty).
\item $u_{\sigma_d}$: depth-estimator per-patch variance, estimated by $8$-pass MC-dropout through Depth-Anything-V2.
\item $u_{\mathrm{pc}}$: photometric inconsistency, computed as the residual between two simulated viewpoints reprojected onto the canonical view.
\end{itemize}

\begin{table}[H]
\centering\footnotesize
\renewcommand{\arraystretch}{1.0}
\setlength{\tabcolsep}{10pt}
\begin{tabular}{lcc}
\hline
Depth-uncertainty proxy & Pearson $\rho(\alpha, -u)$ $\uparrow$ & $p$-value \\
\hline
$u_{\nabla}$ (depth-gradient magnitude) & $+0.61$ & $< 10^{-6}$ \\
$u_{\sigma_d}$ (MC-dropout variance)    & $+0.54$ & $< 10^{-6}$ \\
$u_{\mathrm{pc}}$ (photometric inconsist.) & $+0.47$ & $< 10^{-6}$ \\
\hline
\end{tabular}
\vspace{0.25cm}
\caption{\small Pearson correlation between the learned opacity $\alpha$
and three depth-uncertainty proxies (sign convention: higher $\alpha$
$\Leftrightarrow$ lower uncertainty). Although $\alpha$ is not directly conditioned on any of these proxies, the opacity-weighted GST loss in Eq. 21 in the main paper induces a strong correlation consistent with the "depth-aware confidence" interpretation.}
\label{tab:supp_alpha_corr}
\end{table}

\subsection{Action-Chunk Horizon Sweep on CALVIN}

We evaluate the effect of the action-chunk horizon $H$ on CALVIN ABCD$\rightarrow$D
using the 10\% training subset. The horizon controls how many future actions are predicted
per policy query, creating a trade-off between short-horizon reactivity and long-horizon
temporal consistency.

\begin{table}[H]
\centering
\footnotesize
\renewcommand{\arraystretch}{1.0}
\setlength{\tabcolsep}{8pt}
\begin{tabular}{lcccccc}
\hline
$H$ & 1 & 2 & 3 & 4 & 5 & Avg. Len. \\
\hline
\textbf{10} (default) & \textbf{0.637} & \textbf{0.367} & \textbf{0.291} & \textbf{0.179} & 0.000 & \textbf{1.474} \\
20 & 0.671 & 0.402 & 0.318 & 0.198 & 0.063 & 1.65 \\
50 & 0.598 & 0.331 & 0.237 & 0.122 & 0.030 & 1.32 \\
\hline
\end{tabular}
\vspace{0.25cm}
\caption{\small CALVIN ABCD$\rightarrow$D (10\%) performance under varying action-chunk horizon $H$. The default setting $H{=}10$ provides a strong trade-off between performance and efficiency.}
\label{tab:supp_calvin_h}
\end{table}

Increasing the horizon from $H=10$ to $H=20$ improves the average completed sequence
length from 1.474 to 1.65 and raises the completion rate across all 1--5 task thresholds,
showing that longer chunks can help preserve temporal consistency in CALVIN. However,
longer chunks also reduce the frequency at which the policy can react to new observations,
which can be undesirable for closed-loop control and real-world deployment.

At $H=50$, performance drops across all metrics, indicating that very long action chunks
become harder to predict reliably and are more vulnerable to compounding errors. We
therefore use $H=10$ as the default setting in the main experiments because it provides a
strong accuracy--efficiency trade-off while maintaining frequent replanning.

\begin{figure*}[htbp]
    \centering
    \begin{minipage}{0.19\textwidth}
        \includegraphics[width=\linewidth]{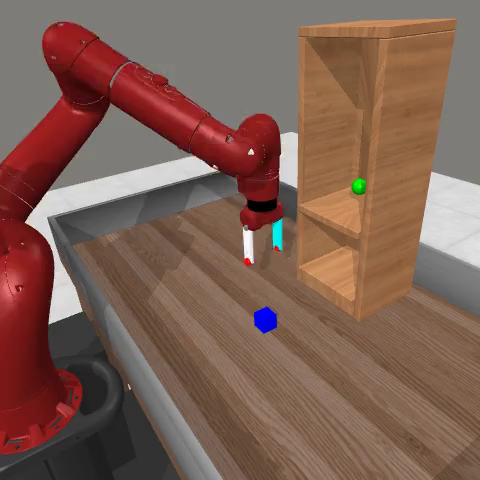}
    \end{minipage}\hfill
    \begin{minipage}{0.19\textwidth}
        \includegraphics[width=\linewidth]{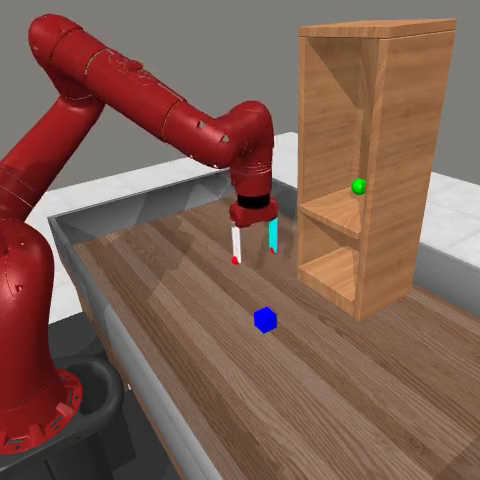}
    \end{minipage}\hfill
    \begin{minipage}{0.19\textwidth}
        \includegraphics[width=\linewidth]{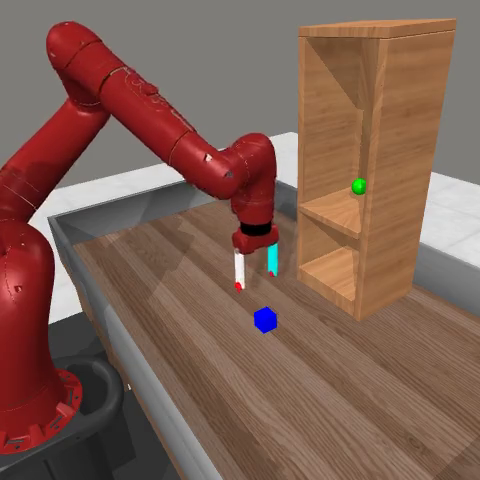}
    \end{minipage}\hfill
    \begin{minipage}{0.19\textwidth}
        \includegraphics[width=\linewidth]{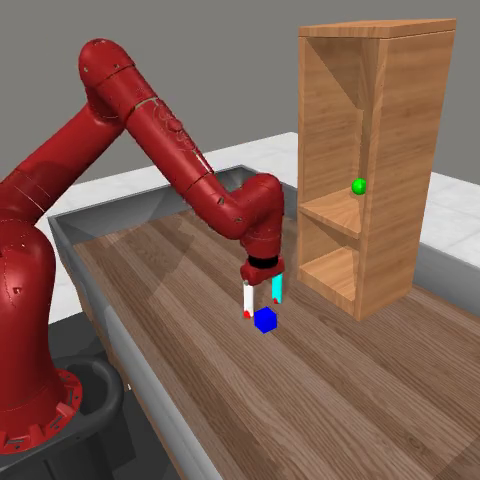}
    \end{minipage}\hfill
    \begin{minipage}{0.19\textwidth}
        \includegraphics[width=\linewidth]{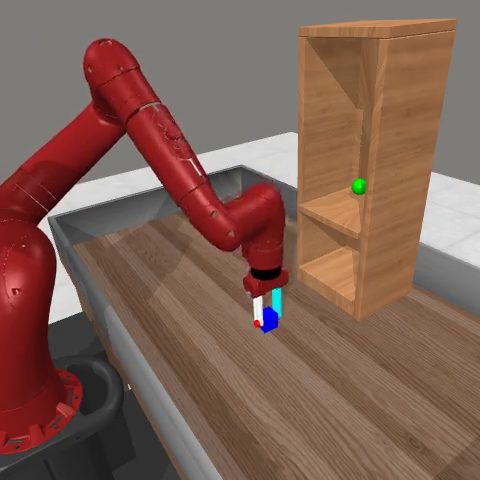}
    \end{minipage}
    
    \vspace{0.2cm} 
    
    \begin{minipage}{0.19\textwidth}
        \includegraphics[width=\linewidth]{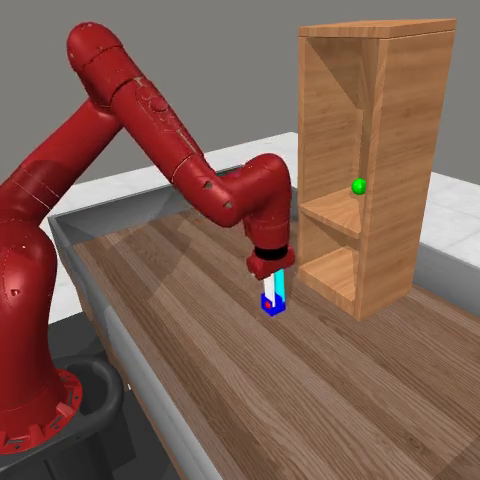}
    \end{minipage}\hfill
    \begin{minipage}{0.19\textwidth}
        \includegraphics[width=\linewidth]{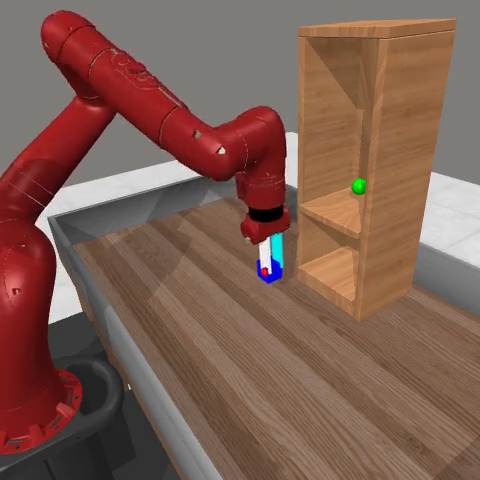}
    \end{minipage}\hfill
    \begin{minipage}{0.19\textwidth}
        \includegraphics[width=\linewidth]{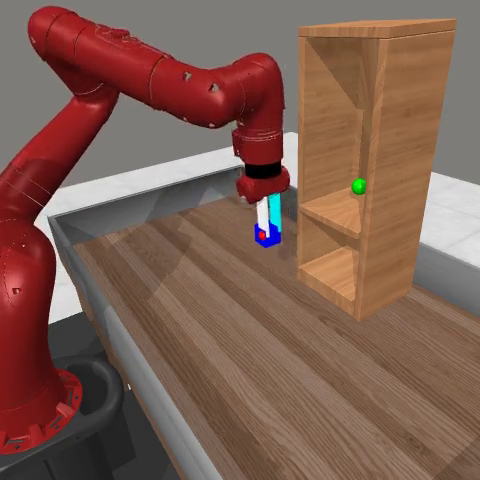}
    \end{minipage}\hfill
    \begin{minipage}{0.19\textwidth}
        \includegraphics[width=\linewidth]{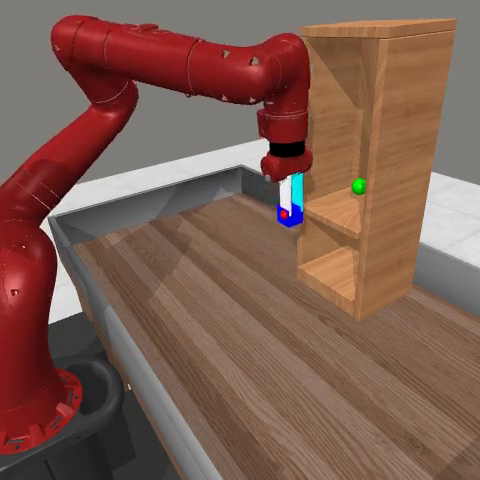}
    \end{minipage}\hfill
    \begin{minipage}{0.19\textwidth}
        \includegraphics[width=\linewidth]{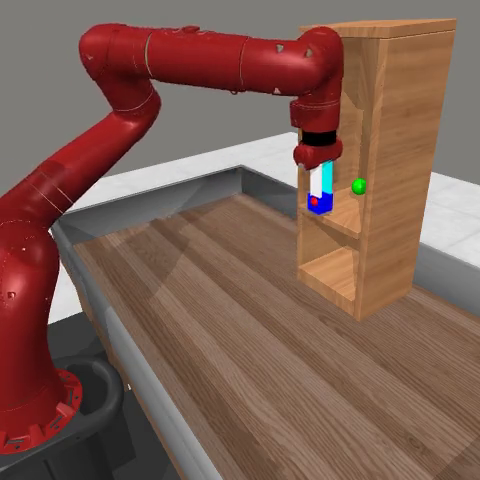}
    \end{minipage}
    \vspace{0.25cm}
    \caption{Instruction(very hard): "Pick and place a puck onto a shelf." Sequential progression of successful task execution in the Meta-world environment.}
    \label{fig:meta_world_sequence}
\end{figure*}
\subsection{Meta-World Successful Frames}

Figure~\ref{fig:meta_world_sequence} shows a representative successful rollout on a
Very Hard Meta-World task. The sequence illustrates how GaussVLA approaches the object,
estimates the target shelf location, grasps the puck, and completes the placement while
maintaining stable control throughout the trajectory.

This example complements the quantitative Meta-World results by showing that GaussVLA
can solve a spatially demanding manipulation task requiring object localization, grasping,
and precise placement under the Very Hard setting.

\end{document}